\documentclass{article}
\usepackage{jfrExamplee}
\usepackage{graphicx}
\usepackage{apalike}
\usepackage{setspace}

\usepackage[ruled,vlined,linesnumbered]{algorithm2e}
\usepackage{hyperref}
\usepackage{siunitx}

\usepackage{amsmath}
\usepackage{gensymb}
\usepackage{amssymb}
\usepackage{makecell}

\usepackage{color}
\usepackage{soul}

\newcommand{\ie}{\textit{i}.\textit{e}.}
\newcommand{\eg}{\textit{e}.\textit{g}.}

\title{LiDAR-based Quadrotor for Slope Inspection in Dense Vegetation}

\author{
Wenyi Liu \\
Department of Mechanical Engineering \\
University of Hong Kong \\
Pokfulam, Hong Kong \\
\texttt{liuwenyi@connect.hku.hk} \\
\And
Yunfan Ren \\
Department of Mechanical Engineering \\
University of Hong Kong \\
Pokfulam, Hong Kong \\
\texttt{renyf@connect.hku.hk} \\
\AND
Rui Guo \\
Department of Mechanical Engineering \\
University of Hong Kong \\
Pokfulam, Hong Kong \\
\texttt{u3619035@connect.hku.hk} \\
\And
Vickie W. W. Kong \\
Geotechnical Engineering Office \\
Civil and Engineering Development Department \\
The Government of Hong Kong SAR \\
Hong Kong, China \\
\texttt{vickiewwkong@cedd.gov.hk} \\
\And
Anthony S. P. Hung \\
Geotechnical Engineering Office \\
Civil and Engineering Development Department \\
The Government of Hong Kong SAR\\
Hong Kong, China \\
\texttt{asphung@cedd.gov.hk} \\
\And
Fangcheng Zhu \\
Department of Mechanical Engineering \\
University of Hong Kong \\
Pokfulam, Hong Kong \\
\texttt{zhufc@connect.hku.hk} \\
\And
Yixi Cai \\
Department of Mechanical Engineering \\
University of Hong Kong \\
Pokfulam, Hong Kong \\
\texttt{yixicai@connect.hku.hk} \\
\And
Yuying Zou \\
Department of Mechanical Engineering \\
University of Hong Kong \\
Pokfulam, Hong Kong \\
\texttt{zyycici@connect.hku.hk} \\
\And
Fu Zhang \\
Department of Mechanical Engineering \\
University of Hong Kong \\
Pokfulam, Hong Kong \\
\texttt{fuzhang@hku.hk} \\
}

\begin{document}

\maketitle

\begin{abstract}

This work presents a LiDAR-based quadrotor system for slope inspection in dense vegetation environments. The primary objective of slope inspection is to access man-made structures erected on hillsides to ascertain the need for slope maintenance, which plays a critical role in ensuring the safety and daily lives of residents. Cities like Hong Kong are vulnerable to climate hazards such as extreme rainfall and typhoons, which often result in landslides. To mitigate the landslide risks, the Geotechnical Engineering Office (GEO) of the Civil Engineering and Development Department (CEDD) has constructed steel flexible debris-resisting barriers on vulnerable natural catchments to protect residents from the danger of landslide debris flow. However, it is necessary to carry out regular inspections to identify any anomalies, such as accumulation of debris behind barriers or severe corrosion of the steel components, which may affect the proper functioning of the barriers. Traditional manual inspection methods face challenges and high costs due to steep terrain and dense vegetation. Compared to manual inspection, unmanned aerial vehicles (UAVs) equipped with LiDAR sensors and cameras have advantages such as maneuverability in complex terrain, access to narrow areas and high spots, and the ability to collect detailed topographic and obstacle data, making them more suitable for slope inspection. However, conducting slope inspections using UAVs in dense vegetation poses significant challenges. First, in terms of hardware, the overall design of the UAV must carefully consider its maneuverability in narrow spaces, flight time, and the types of onboard sensors required for effective inspection. Second, regarding software, navigation algorithms need to be designed to enable obstacle avoidance flight in dense vegetation environments. While research and commercial solutions exist for bridge inspection, power line inspection, and assisted obstacle avoidance, they all have their limitations. To overcome these challenges, we develop a LiDAR-based quadrotor, accompanied by a comprehensive software system comprising localization, mapping, planning, and control algorithms. The goal is to deploy our quadrotor in field environments to achieve efficient slope inspection. To assess the feasibility of our hardware and software system, we conduct functional tests on our quadrotor in non-operational scenarios. Subsequently, invited by CEDD to develop UAVs for visual inspection of flexible debris-resisting barriers and other geotechnical features, we deploy our quadrotor in six field environments, including five flexible debris-resisting barriers located in dense vegetation and one slope that experienced a landslide caused by the rainstorm. In all these six field tests, our quadrotor effectively accomplishes the assigned inspection tasks. Additionally, we conduct comparative experiments between our quadrotor and the advanced commercial drone DJI Mavic 3 in terms of assisted obstacle avoidance flight. These experiments demonstrated the superiority of our quadrotor in terms of dynamic obstacle avoidance and maneuvering capabilities in narrow areas, as well as its applicability in slope inspection.

Keyword: slope inspection, dense vegetation, LiDAR-based quadrotor

\end{abstract}

\section{Introduction}

\begin{figure}[h]
    \centering
    \includegraphics[width=0.95\textwidth]{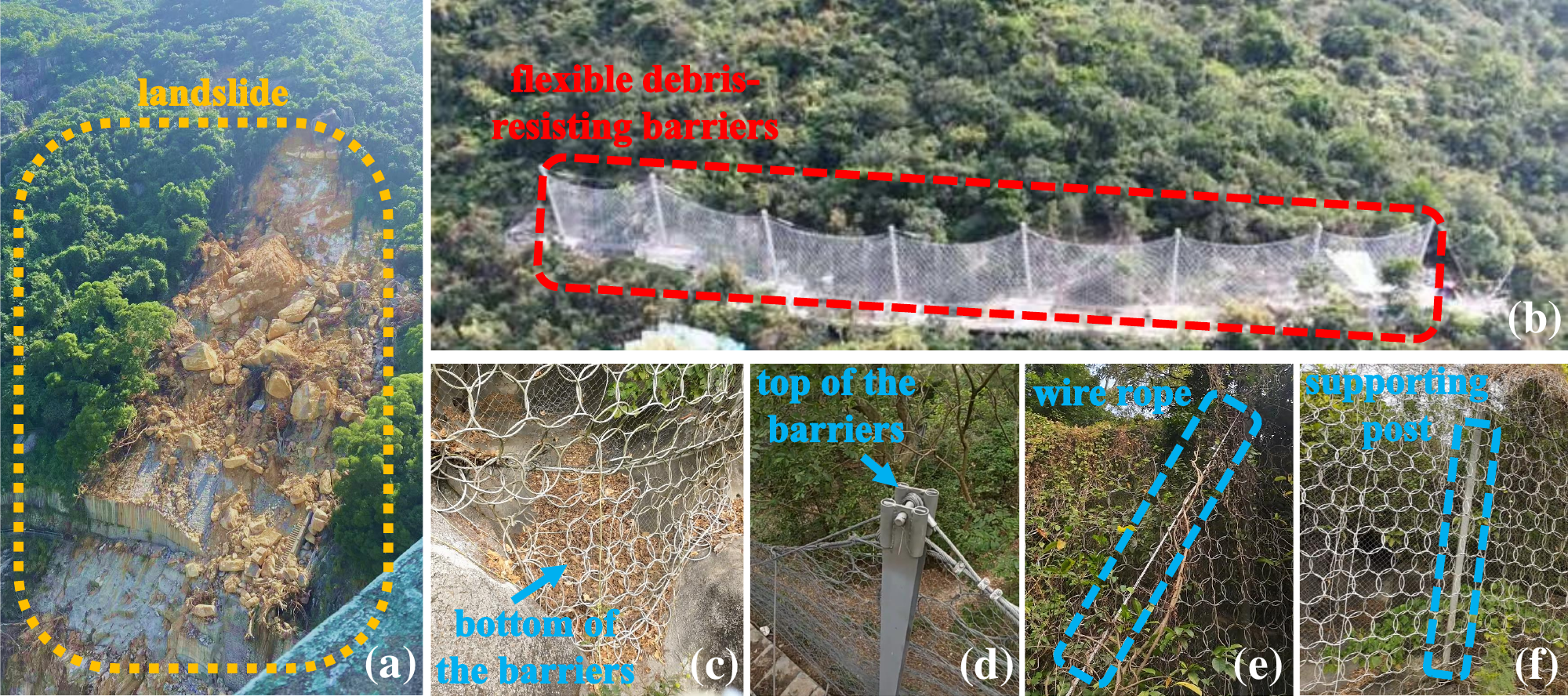}
    \caption{(a) A slope in Hong Kong that suffered a significant landslide in September 2023. (b) Flexible debris-resisting barriers constructed on a slope to stop landslides. (c)-(f) Inspection targets of the flexible debris-resisting barriers.}
    \label{fig:env}
\end{figure}

Slope inspection is a crucial task performed in complex and unknown environments to inspect man-made structures, including steel flexible debris-resisting barriers, erected on hillsides to ascertain the need for slope maintenance. It plays a vital role in ensuring the safety and daily lives of residents. In regions like Hong Kong, which are characterized by coastal and mountainous terrain, climate hazards like extreme rainfall and typhoons pose significant risks, often leading to events such as landslides. On average, there are about three hundred landslides occur in Hong Kong each year and some landslides can have severe consequences as shown in Fig. \ref{fig:env}(a). The Geotechnical Engineering Office (GEO) of the Civil Engineering and Development Department (CEDD) is tasked by the Government of Hong Kong Special Administrative Region to manage the landslide risk for Hong Kong. The GEO has implemented a Slope Safety System to achieve the mandate given. One of the key strategies in the Slope Safety System is to adopt an engineering approach to stabilize existing substandard man-made slopes and mitigate landslide risk from natural hillsides.  As regards the latter, it has been the practice to erect flexible debris-resisting barriers at strategic locations to intercept landslide debris coming from uphill and hence protect the public and infrastructure downhill, as shown in Fig. \ref{fig:env}(b).

Flexible debris-resisting barriers are usually erected in the mid-slope of the hillside, which makes it difficult to access and the environment can change over time (\eg, the growth of vegetation and trees). However, regular inspection of the flexible debris-resisting barriers is important to ascertain the proper functioning of these barriers, particularly for the steel flexible barrier. Any anomalies, such as accumulation of debris behind barriers (Fig. \ref{fig:env}(c)) or severe corrosion of the steel components, may call for subsequent maintenance and repair works. These inspections encompass a comprehensive assessment of various aspects related to the flexible debris-resisting barriers, including the steel components, such as the wire ropes positioned atop the barriers (Fig. \ref{fig:env}(d)), the inclined wire ropes (Fig. \ref{fig:env}(e)) and the supporting posts (Fig. \ref{fig:env}(f)), to ensure the stability of the barriers. Diligently conducting these inspections ensures that the flexible barrier can effectively mitigate the impact of debris when it becomes necessary.

\begin{figure}[h]
    \centering
    \includegraphics[width=0.92\textwidth]{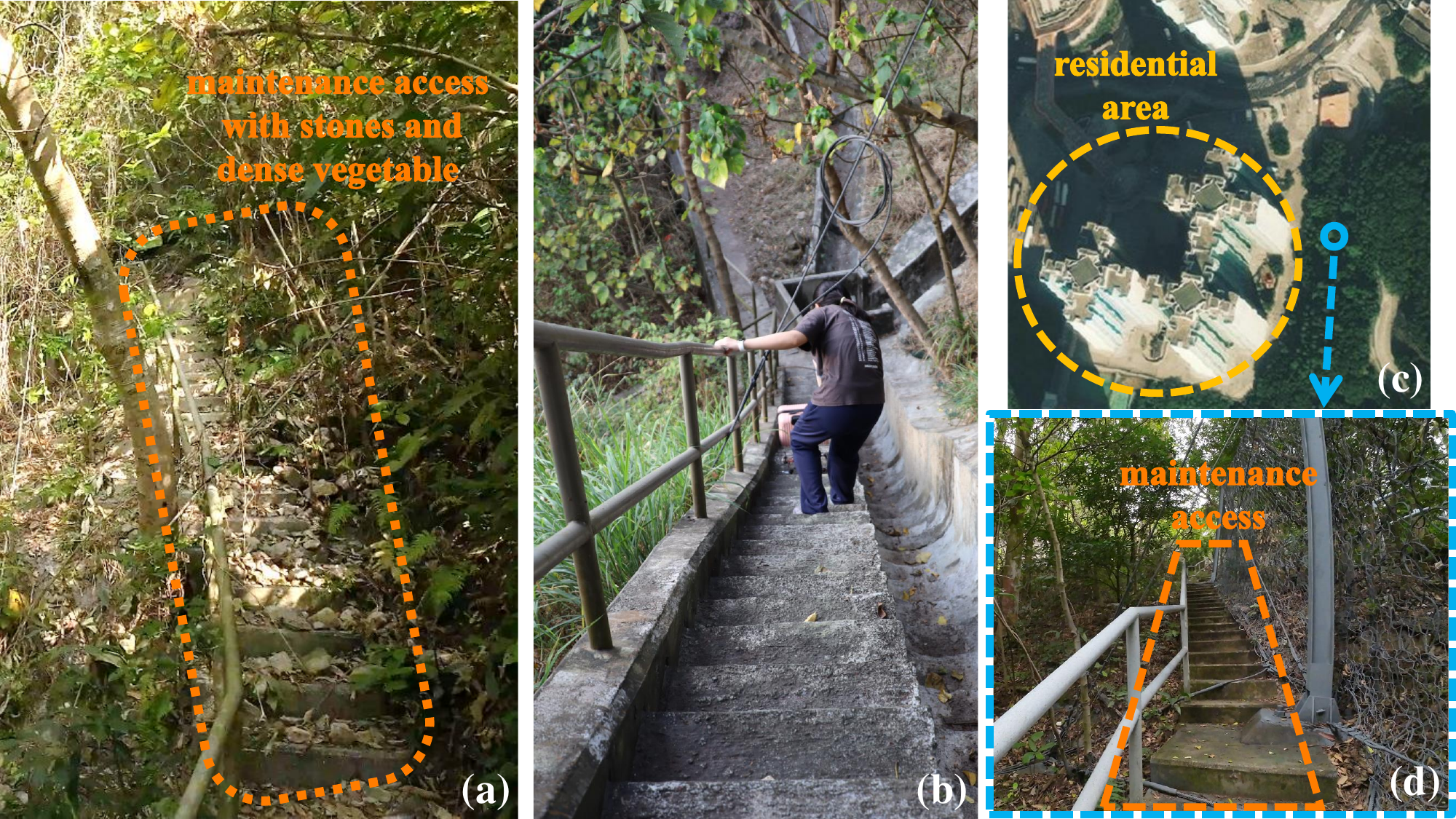}
    \caption{(a) Inspection maintenance access with stones and dense vegetation. (b) Personnel conducts an inspection on the maintenance access. (c-d) A maintenance access near residential areas.}
    \label{fig:access}
\end{figure}

Currently, the maintenance agent of the debris-resisting barrier employs a manual inspection approach. Maintenance access is constructed along the flexible debris-resisting barriers on the slope to provide safe access for the inspecting personnel, as depicted in Fig \ref{fig:access}. However, the location of these barriers necessitates the construction of long maintenance access to access these barriers from public roads. As shown in Fig. \ref{fig:access}(a) and Fig. \ref{fig:access}(b), the lengthy maintenance access, combined with the steep terrain, leads to high construction costs and imposes an increased workload on the inspecting personnel. Additionally, dense vegetation and hostile environments (\eg, steep rugged stairs, insects) are not suitable for inspecting personnel to stay. Furthermore, providing access in proximity to residential areas (see Fig. \ref{fig:access}(c)) is not welcome for the reasons of privacy and crime prevention. 

Compared to manual inspection, unmanned aerial vehicles (UAVs)-based inspection possesses unique advantages and potential applications in the field of slope inspections. Firstly, UAVs offer the capability to navigate complex terrains, covering wide areas and accessing areas that are difficult for humans or ground mobile robots to reach. They also provide rich terrain and obstacle data through onboard sensors such as LiDAR and cameras. Additionally, the deployment of UAVs reduces the need for human labor, mitigates operational risks, and enhances work efficiency. Moreover, the utilization of UAVs can potentially eliminate the necessity for access construction and maintenance, resulting in substantial cost savings and reduced criminal activities.

However, applying UAVs to slope inspections in dense vegetation presents several hardware and software challenges. In terms of hardware challenges, it is necessary to design the UAV's mechanical structure systematically, considering factors such as maneuverability in narrow areas, flight time, and the types of onboard sensors required for effective inspections. For software challenges, localization, mapping, and planning and control in navigation algorithms play a crucial role. Firstly, in terms of \textbf{localization}, traditional GPS-based methods become unreliable in dense vegetation environments. UAVs need to rely on onboard sensors such as LiDAR and cameras to achieve accurate localization. Secondly, for \textbf{mapping}, It is necessary to use data from onboard sensors to rapidly and accurately update information about thin objects in the map, such as thin tree branches, wire ropes, and fine nets on barriers. Finally, for \textbf{planning and control}, UAVs must effectively mitigate wind disturbances while responding smoothly and quickly to avoid dynamic objects, enabling safe navigation amidst swaying tree branches and other natural disturbances.

Many works \cite{castelar2024lidar,chen2019uav,jordan2018state,li2023design,winkvist2013towards,nikolic2013uav} deploy UAVs for inspection tasks on structures such as bridges, power lines, and buildings. These UAVs are capable of conducting inspections in areas away from obstacles. Limitations such as the larger size of the aircraft or the lack of comprehensive navigation algorithms render these approaches unsuitable for narrow-area inspection. To address the challenge of narrow area inspections, some studies \cite{briod2014collision,salaan2018close} and commercial UAVs such as Elios 3 \cite{elios3} and Dronut X1 Pro \cite{dronut} design protective enclosures around the UAV's propellers, ensuring flight stability during collisions with regular obstacles. Nevertheless, in dense vegetation environments, the presence of thin objects such as thin tree branches and vines can become entangled in the UAV's rotors, significantly impacting flight stability. Additionally, some works \cite{jimenez2015aerial,ikeda2019stable} incorporate camera-equipped robotic arms on UAVs to reach into narrow areas. However, the limited length of these arms severely restricts the inspection range in narrow areas, making it impractical to perform inspections in dense vegetation. Currently, commercial UAVs such as DJI Mavic 3 \cite{mavic3} and Skydio 2 plus \cite{skydio2+}, equipped with multiple sensors and employing navigation algorithms for obstacle avoidance in sparser environments, demonstrate practicality. However, due to the low accuracy of visual measurements and the paramount concern for safety, their obstacle avoidance functionality restricts UAVs from flying in narrow areas. As a result, these platforms are unsuitable for inspection tasks in dense vegetation scenarios.

\begin{figure}[h]
    \centering
    \includegraphics[width=0.92\textwidth]{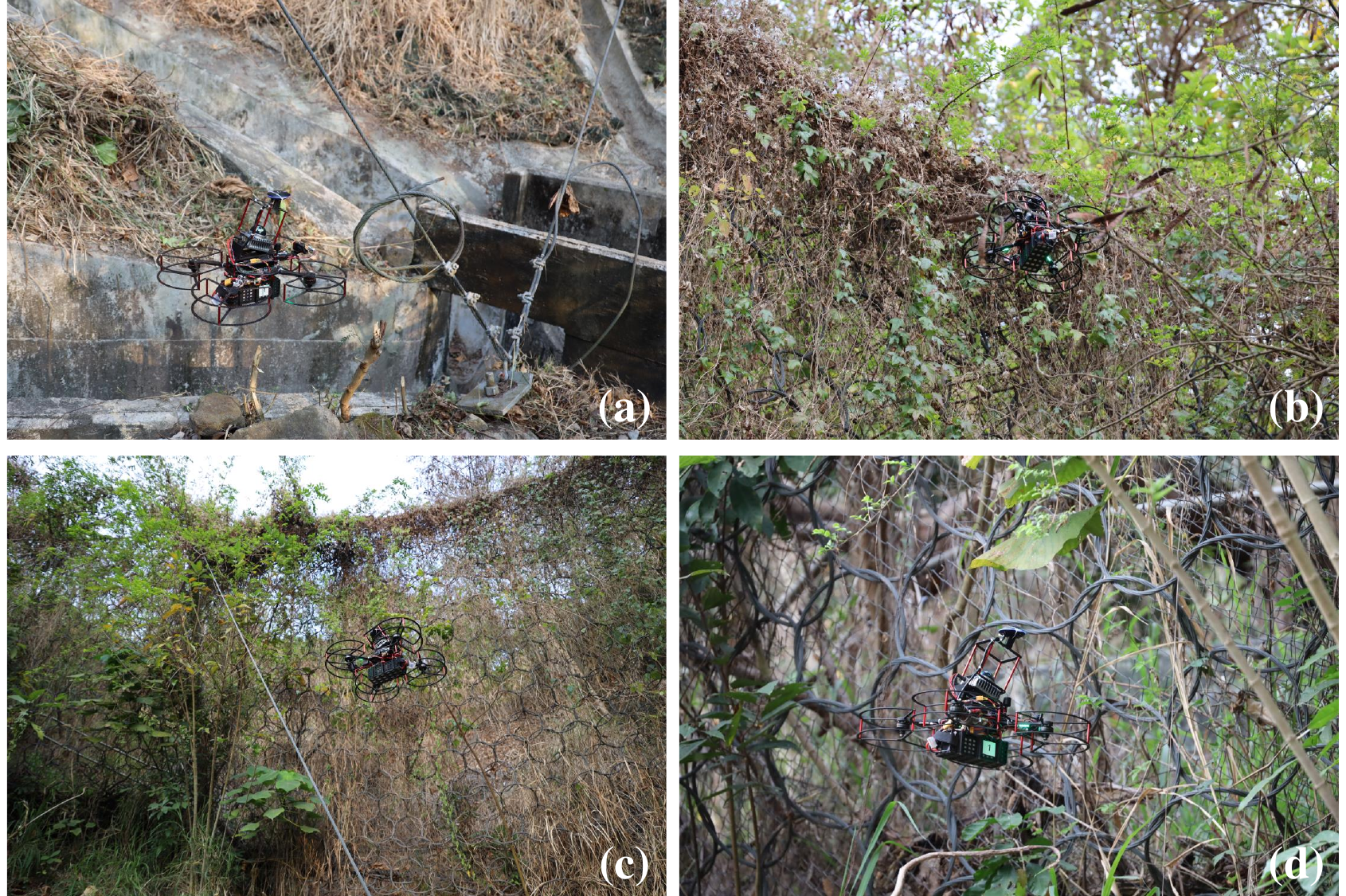}
    \caption{Our quadrotor performs slope inspection in field environments.}
    \label{fig:field_uav}
\end{figure}

To address the challenges of slope inspection in dense vegetation, we propose a comprehensive solution involving UAV hardware and software design. Considering that LiDAR sensors can directly provide high-precision 3D point clouds compared to visual sensors, enabling the reconstruction of barrier geometry during slope inspections, we develop a quadrotor equipped with a LiDAR sensor in Sec. \ref{sec:hardware}. Our quadrotor possesses a wheelbase of \SI{320}{mm} and overall dimensions measuring \SI{422}{mm} in length, \SI{422}{mm} in width, and \SI{260}{mm} in height, with a total mass of \SI{2.1}{kg}, allowing it to navigate through narrow areas. It also provides a maximum flight time of 12 minutes. Regarding the software, we systematically design the localization, mapping, planning, and control modules based on our prior works, as described in Sec.\ref{sec:software}. For the localization module (Sec. \ref{sec:lio}), we employ FAST-LIO2 \cite{xu2022fast}, a robust LiDAR-inertial odometry framework capable of operating in complex environments without relying on GPS signals. In the mapping module (Sec. \ref{sec:map}), we enhance ROG-Map \cite{ren2023rog}, which constructs real-time high-resolution sliding window grid maps. By incorporating three enhancement techniques, namely the Unknown Grid Cells Inflation, Infinite Points Ray Casting and Incremental Frontier Update, our quadrotor can effectively identify the correct gird occupancy states in case of no LiDAR returns or with returns caused by thin objects, allowing for efficient avoidance of potential obstacles in dense vegetation. For the planning and control modules (Sec. \ref{sec:ipc}), we re-design the frontend and backend of the integrated planning and control framework, IPC \cite{liu2023integrated}, to incorporate the assisted obstacle avoidance flight function. This augmentation assists the pilot in avoiding static and dynamic obstacles, ensuring safe navigation during inspections. As shown in Fig. \ref{fig:field_uav}, our LiDAR-based quadrotor can be deployed in field environments, enabling efficient slope inspections, improving safety, and reducing the workload of personnel involved in the inspection task.

To assess the feasibility of our hardware and software system, we conduct functional tests on our quadrotor in non-operational scenarios. Subsequently, invited by CEDD to develop UAV for visual inspection of flexible debris-resisting barriers and other geotechnical features, we deploy our quadrotor in six field environments, including five flexible debris-resisting barriers located in dense vegetation and one slope that experienced a landslide caused by the rainstorm. In all these six field tests, our quadrotor effectively accomplishes the assigned inspection tasks. Additionally, we conduct comparative experiments between our quadrotor and the advanced commercial drone DJI Mavic 3 in terms of assisted obstacle avoidance flight. These experiments demonstrated the superiority of our quadrotor in terms of dynamic obstacle avoidance and maneuvering capabilities in narrow areas, as well as its applicability in slope inspection.

In the subsequent sections, we will provide a detailed description of the hardware structure in Sec. \ref{sec:hardware}, including the rationale behind selecting the LiDAR sensor, the mechanical structure, the electrical system, and essential parameters such as flight time. Following that, we will introduce the software structure in Sec. \ref{sec:software}, which comprises localization (Sec. \ref{sec:lio}), mapping (Sec. \ref{sec:map}), and planning and control (Sec. \ref{sec:ipc}). Each section will extensively cover related work, the advantages of our approach, and specific technical details. Next, we will present the experiment in Sec \ref{sec:experiments}. In this section, we will introduce the functional tests in non-operational scenarios (Sec. \ref{sec:indoor_test}), the six field tests (Sec. \ref{sec:field_test}), and the benchmark experiments (Sec. \ref{sec:benchmark}) comparing our quadrotor with the advanced commercial drone, DJI Mavic 3, further validating our quadrotor's suitability for slope inspection in dense vegetation. Finally, we will conclude our work in Sec. \ref{sec:conclusion}, summarizing the key contributions and highlighting the significance of our proposed solution. More details can be found in the attached video\footnote{\href{https://youtu.be/CE92FNn2eDY}{https://youtu.be/CE92FNn2eDY}}.

\section{Hardware Structure}
\label{sec:hardware}

Currently, numerous drones utilize cameras for environmental perception. While cameras offer cost-effectiveness and provide rich visual information, they encounter significant limitations and challenges in slope inspections. Firstly, cameras are sensitive to lighting conditions and weather. In environments with dense vegetation and sunny weather, the lighting conditions can be complex, and the camera may be affected by sunlight or dense shadow, resulting in overexposure or underexposure and degrading navigation robustness. Secondly, camera-based visual navigation has a limited mapping resolution, making it difficult to sense thin objects in dense vegetation, such as thin tree branches, wire ropes, and fine nets on barriers. Moreover, visual navigation has limited mapping accuracy and large mapping noises due to complex lighting or environment contents, making it challenging for camera-based drones to safely navigate through narrow spaces in dense vegetation required by slope inspections.

In contrast, LiDAR exhibits significant advantages in slope inspection. Firstly, LiDAR actively emits laser beams and is not affected by lighting conditions. It maintains stable performance even in complex lighting or low-lighting scenarios within dense vegetation. Secondly, the high measurement accuracy of LiDAR facilitates the creation of high-accuracy, high-resolution maps, enabling LiDAR-based drones to navigate safely through narrow spaces in dense vegetation while avoiding thin tree branches, wire ropes, fine nets etc. during slope inspections. Moreover, LiDAR also provides precise 3D point cloud data for a comprehensive assessment of the flexible debris-resisting barriers on the slopes. Therefore, we select LiDAR as the primary sensor for our quadrotor. Considering the payload limitations of the quadrotor, we select the Livox Mid-360 LiDAR\cite{mid360}, which weighs only \SI{265}{g}. This LiDAR employs a non-repetitive scanning approach, accumulating data over time to generate dense point cloud maps. Besides, it features a 360-degree horizontal field of view and a 59-degree vertical field of view, enabling the perception of a wide range of scenes.

\begin{figure}[h]
    \centering
    \includegraphics[width=\textwidth]{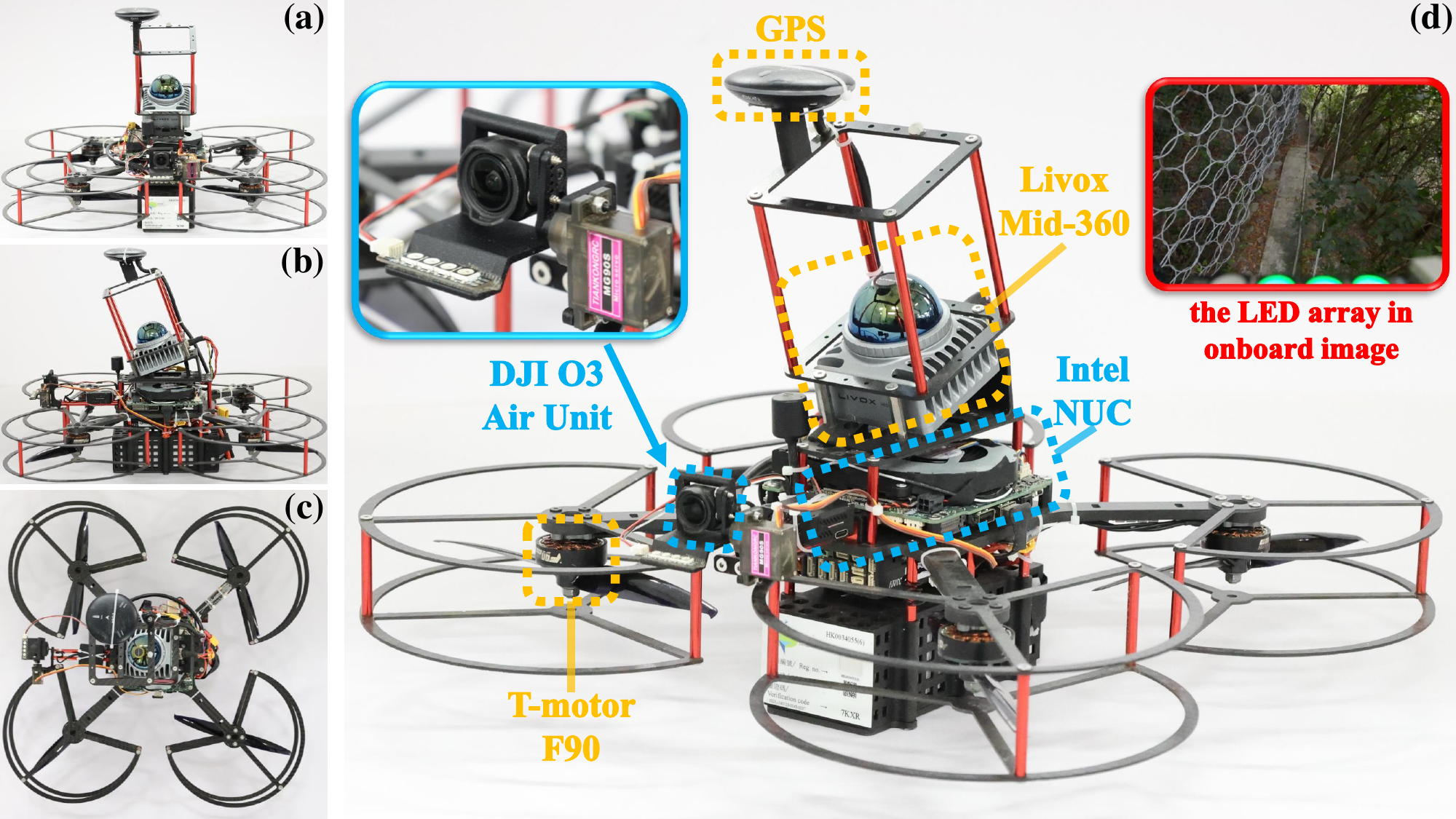}
    \caption{Different views of our LiDAR-based quadrotor.}
    \label{fig:uav}
\end{figure}

\begin{table}[h]
    \centering
    \small
    \caption{Device Information of our In-House Developed Quadrotor.}
    \label{tab:hardware}
    \begin{tabular}{|c|c|c|}
        \hline
        Device & Description & Weight (g) \\
        \hline
        ESC & T-motor F60A 8S 4IN1 & 15.3 \\
        \hline
        Motor & T-motor F90 KV1300 & 41.8 \\
        \hline
        Propeller & GEMFAN Flash 7042 & 5.48 \\
        \hline
        Receiver & RadioLink R12DSM & 2.5 \\
        \hline
        GPS sensor & CUAV NEO 3 with NEO-M9N & 33 \\
        \hline
        Flight Controller & CUAV Nora+ & 91 \\
        \hline
        Battery & ACE 6S-5300mAh-30C lithium battery & 664 \\
        \hline
        Onboard Computer & Intel NUC with Intel i7-1260P CPU & 270 \\
        \hline
        LiDAR & Livox Mid-360 & 265 \\
        \hline
        FPV Camera & DJI O3 Air Unit & 36.4 \\
        \hline
        Goggle glasses & DJI Goggles 2 & 290 \\
        \hline
        Remote Controller & RadioLink AT9S PRO & 980 \\
        \hline
    \end{tabular}
\end{table}

Then we present a comprehensive description of the hardware configuration implemented on our LiDAR-based quadrotor platform, as shown in Table \ref{tab:hardware} and Fig. \ref{fig:uav}. Firstly, our quadrotor utilizes an Intel NUC mini-computer with an Intel i7-1260P CPU chip, capable of operating at a high frequency of \SI{4.7}{GHz}. This onboard computer provides substantial computational power for real-time processing tasks. To enable real-time observation of the flight process by the pilot and capture photos and videos of specific areas, we use an FPV camera DJI O3 Air Unit\footnote{\href{https://www.dji.com/o3-air-unit}{https://www.dji.com/o3-air-unit}}, a high-definition digital video transmission system, with the DJI Goggles 2\footnote{\href{https://www.dji.com/goggles-2}{https://www.dji.com/goggles-2}}. This video transmission system offers an impressively low latency of \SI{40}{ms}, providing timely feedback on the surrounding environment for the pilot. Moreover, it supports recording \SI{4}{K} videos at a high frame rate of \SI{120}{Hz}, enabling in-depth post-analysis after the flights. In order to enlarge the Field of View of the FPV camera, so as to observe both the top and bottom of the flexible debris-resisting barriers on the slope, we install the camera on a pitch-axis gimbal. The pitch angle of the gimbal is commanded by the remote controller in real time during the flight. To maximize the flight time, which is a critical consideration, our quadrotor is equipped with 7-inch propellers and a high-capacity 6S-5300mAh battery. We use an LED array to keep track of the quadrotor's battery level in real-time. The LED array consists of four LEDs, indicating the battery percentage. They display green for battery percentage over \SI{40}{\%}, red for battery percentage between \SI{25}{\%} and \SI{40}{\%}, and flash red for battery percentage below \SI{25}{\%}. The LED arrays are installed in front of the camera and their status is visible in the streamed video, so the remote operator can initiate a return flight before the battery is out. To enhance the quadrotor safety, carbon fiber propeller guards are meticulously designed and installed to minimize the risk of propeller-related accidents. These guards protect both the quadrotor and the pilot, ensuring safe and reliable operation.

After these devices are integrated into the airframe composed of carbon plates and aluminum columns, we conduct tests on the developed quadrotor. The quadrotor features a motor-to-motor distance (\ie, wheelbase) of \SI{320}{mm} and overall dimensions of \SI{422}{mm} in length, \SI{422}{mm} in width, and \SI{260}{mm} in height. With a total mass of \SI{2.1}{kg}, the quadrotor achieves a thrust-to-weight ratio of 3, ensuring efficient and stable flight performance. Additionally, the maximum flight time is measured to be 12 minutes.

\section{Software Structure}
\label{sec:software}

\begin{figure}[h]
    \centering
    \includegraphics[width=0.9\textwidth]{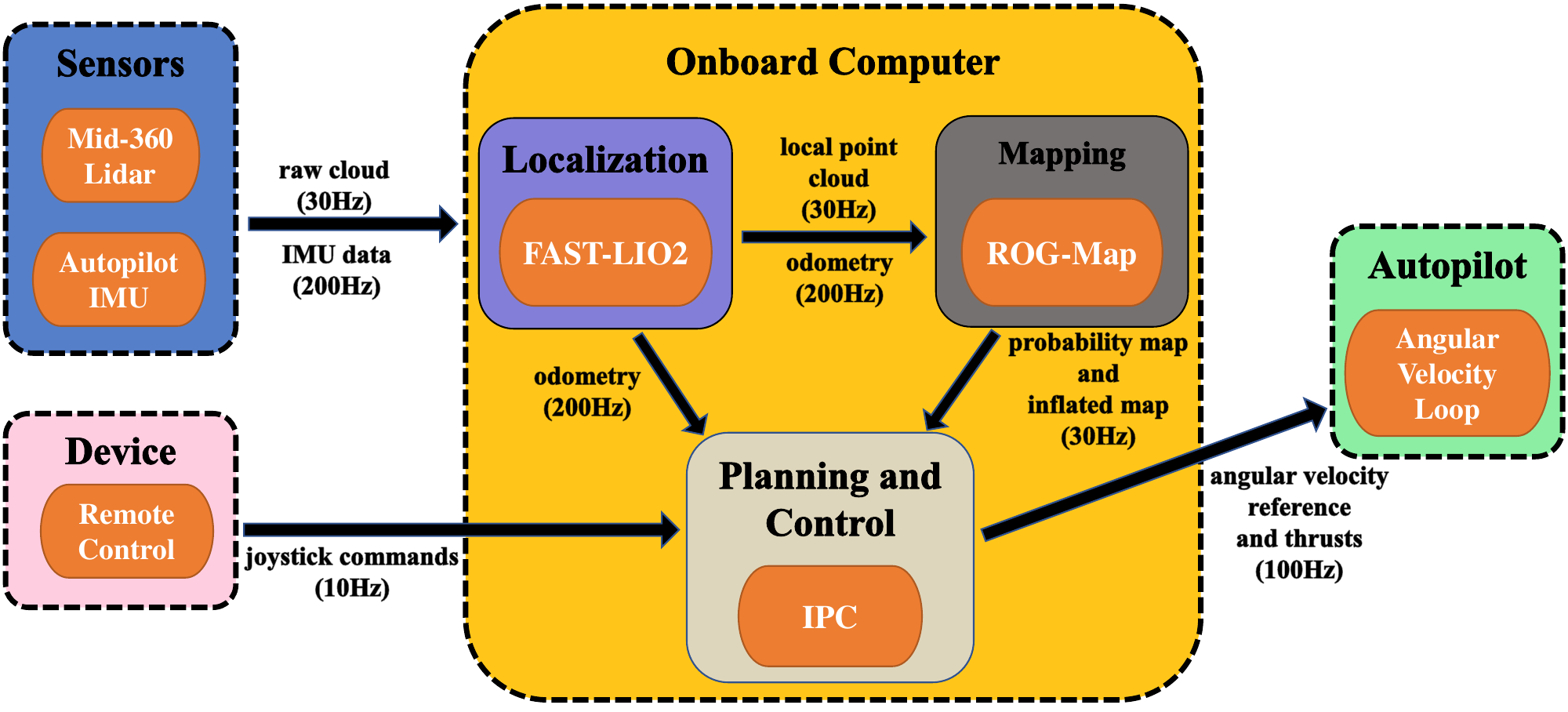}
    \caption{The software structure of our quadrotor.}
    \label{fig:software}
\end{figure}

The software structure of our quadrotor is illustrated in Fig. \ref{fig:software}. All navigation modules run in real-time on the onboard computer. Our localization module employs FAST-LIO2 \cite{xu2022fast}, which utilizes an iterative error-state Kalman filter and an incremental kd-Tree (\ie, ikd-Tree) to provide the quadrotor's odometry and local point cloud in the world frame. This module takes as input the received LiDAR raw data and IMU data for accurate estimation. The mapping module, an extension of our previous work, ROG-Map \cite{ren2023rog}, incorporates additional features including Unknown Grid Cells Inflation (Sec. \ref{sec:unk_inf}), Infinite Points Ray Casting (Sec. \ref{sec:inf_point}), and Incremental Frontiers Updates (Sec. \ref{sec:map_frontier}). These enhancements are built upon the foundations of Map Sliding and Incremental Inflation implemented in ROG-Map. The mapping module generates a local occupancy grid map (OGM) for obstacle avoidance. Our planning and control module is based on our previous work, IPC \cite{liu2023integrated}, where the frontend is redesigned for assisted obstacle avoidance flight. The IPC directly generates angular velocity references and throttle commands for the quadrotor based on local goal specified in real-time by joystick, odometry, probability map and inflated map, ensuring assisted obstacle avoidance flight in dense vegetation environments. In the subsequent sections, we will provide a detailed description of the localization (Sec. \ref{sec:lio}), mapping (Sec. \ref{sec:map}), and planning and control modules (Sec. \ref{sec:ipc}).

\subsection{Localization}
\label{sec:lio}

In slope inspection, the localization module plays a crucial role by providing real-time state estimation (position, velocity, and attitude) to the quadrotor's controller. Additionally, it transforms the LiDAR points in the local frame into the world frame to construct subsequent navigation maps. GPS is incapable of perceiving environmental obstacles, and its signals are weakened by tree canopies and sloping terrain. Meanwhile, cameras are susceptible to rapidly-changing lighting conditions in dense vegetation, incapable of mapping thin obstacles, such as tree branches, vines, wire ropes, and fine nets, that are abundant in slope environments. In contrast, LiDAR emerges as a more suitable alternative for slope inspection, as it remains unaffected by factors such as lighting variations and terrain irregularities. It can also map thin obstacles reliably as shown in \cite{kong2021avoiding}.

In LiDAR odometry, optimization-based methods such as LOAM \cite{zhang2014loam}, LeGO-LOAM\cite{shan2018lego}, LIOM \cite{ye2019tightly}, LIO-SAM \cite{shan2020lio}, and LILIOM \cite{li2021towards} tend to be computationally expensive and exhibit limited robustness in dense vegetation environment. In contrast, our previous work FAST-LIO2 \cite{xu2022fast} is a lightweight LiDAR-inertial Odometry (LIO) system. FAST-LIO2 utilizes an iterative extended Kalman filter, tightly coupling high-speed measurements to eliminate LiDAR scan drift and enhance robustness against fast motions. Additionally, FAST-LIO2 develops an incremental kd-Tree \cite{cai2021ikd} for fast and efficient scan-to-map registration. Due to its high computational efficiency and robustness, we directly employ FAST-LIO2 as the localization module in our software.

In this work, our adapted FAST-LIO2 utilizes the \SI{30}{Hz} raw point cloud from LiDAR and the \SI{200}{Hz} IMU. By performing IMU pre-integration, FAST-LIO2 can provide low-latency state estimation at a frequency of \SI{200}{Hz} with a latency of less than \SI{1}{ms}. Additionally, it generates world-frame registered point clouds at a scan rate of \SI{30}{Hz}, producing about 200,000 points per second with a latency of less than \SI{10}{ms}.

\subsection{Mapping}
\label{sec:map}

In slope inspection, the mapping module is responsible for constructing high-resolution navigation maps based on the odometry and point cloud data provided by the localization module. These maps accurately represent the terrain and obstacles, serving as a reference for the navigation system and enabling the quadrotor to navigate through dense vegetation while effectively avoiding collisions.

As a promising navigation map type for robots, occupancy grid map (OGM) enables the distinction between occupied, free, and unknown areas in the environment through ray casting and probabilistic updates to handle sensor noise and dynamic objects. Existing methods for implementing occupancy maps can be divided into three main streams: octree-based \cite{hornung2013octomap}, hash table-based \cite{niessner2013real}, and uniform grid-based \cite{zhou2020ego}. Octree-based methods require frequent subdivision of more regions, which reduces query and storage efficiency and increases computation time. Hash table-based methods are prone to hash collisions, and resolving these collisions through techniques such as chaining or open addressing introduces additional computational costs and complexity. Uniform-grid-based methods offer the advantage of map update and access complexity of \textit{O}(1), thereby improving computational efficiency. However, they consume a large amount of memory, which is impractical for high-resolution or large-scale maps. To overcome this limitation, our previous work ROG-Map \cite{ren2023rog}, maintains a high-resolution local occupancy grid map centered around the robot through map sliding, limiting the memory consumption.

In ROG-Map, a zero-copy map sliding strategy is utilized to maintain two local maps. The first local map is the probability map, which stores the occupancy probabilities of grid cells within the local map. When receiving a LiDAR scan in the world frame at time $k$, the probability map utilizes Bayesian update \cite{hornung2013octomap,moravec1985high} to fuse the measurements. If the LiDAR point falls in a grid cell, it is considered a \textbf{hit}, while if the LiDAR beam passes through the grid cell, it is considered a \textbf{miss}. Assuming that the map update process is Markovian, we can use equation (\ref{eq:prob}) and the user-defined measurement probabilities $p_{hit}$ and $p_{miss}$ to update the occupancy probability of the grid cell.
\begin{equation}
    \label{eq:prob}
    \begin{aligned}
        P_{1:k}(\mathbf{n}) &= \left[1 + \mathbf{P}\right]^{-1} \\
        \mathbf{P} &= \frac{1-P_{k}(\mathbf{n})}{P_k(\mathbf{n})}\frac{1-P_{1:k-1}(\mathbf{n})}{P_{1:k-1}(\mathbf{n})}\frac{P(\mathbf{n})}{1-P(\mathbf{n})}
    \end{aligned}
\end{equation}
where $P_k(\mathbf n)$ represents the measurement probability of the grid cell $\mathbf{n}$ at time $k$ (\eg, $p_{hit}$ for a \textbf{hit} or $p_{miss}$ for a \textbf{miss}), $P_{1:k-1}(\mathbf n)$ denotes the occupancy probability of grid cell $\mathbf{n}$ given the measurement history up to time $k-1$, serving as the prior probability before fusing the $k$-th measurement, while $P_{1:k}(\mathbf{n})$ signifies the posterior probability after fusing the $k$-th measurement. $P(\mathbf n)$ is a prior probability, which is commonly assumed as $P(\mathbf n) = 0.5$ to indicate that the map has no prior information of the occupancy state (\ie, the occupancy state of all grid cells in the map is unknown).

To reduce computational complexity, probabilities in ROG-Map are transformed using log-odds (\ref{eq:log_odds}):
\begin{equation}
\label{eq:log_odds}
    L_{(\cdot)}(\mathbf n)= \log\left(\frac{P_{(\cdot)}(\mathbf n)}{1-P_{(\cdot)}(\mathbf n)}\right), 
\end{equation}

Thus, equation (\ref{eq:log_odds}) can be rewritten as equation (\ref{eq:prob_odd}), converting the multiplication operations into addition operations.
\begin{equation}
    \label{eq:prob_odd}
    L_{1:k}(\mathbf n) = L_{1:k-1}(\mathbf n) + L_k(\mathbf {n})
\end{equation}
where $L_{1:k}(\mathbf n)$ denotes the log-odds representation of the fused occupancy probability for grid cell $\mathbf n$ up to time $k$.
Based on user-defined thresholds $l_{occ}$ and $l_{free}$, the probability map classifies the grid cell states into three categories: Occupied, Unknown, and Known Free.

The second local map is the inflated map, utilized for robot navigation in configuration space by inflating obstacles. In the inflated map, the grid cell states are categorized as either Inflation or No Inflation. However, unlike the update mechanism of the probability map (\ie, the first map), the inflated map adopts an incremental update mechanism. This mechanism is achieved by maintaining a counter for each grid cell. Specifically, when a grid cell state in the probability map changes from Unknown or Known Free to Occupied, the counter of the grid cell in the inflated map and its inflated neighbors is incremented by 1, thereby setting the status of these grid cells as Inflation. Conversely, when a grid cell state in the probabilistic map changes from Occupied to Unknown or Known Free, the counter of the corresponding grid cell in the inflated map and its inflated neighbors is decremented by 1. If the counter of a grid cell after the decrement is greater than zero, it indicates that the grid cell is still inflated by other Occupied grids and its state remains Inflation. If the counter is zero, this grid cell is in a No Inflation state.

In this work, we introduce three enhancements to ROG-Map specifically tailored for dense vegetation environments. These enhancements include Unknown Grid Cells Inflation, Infinite Points Ray Casting, and Incremental Frontiers Update. ROG-Map only inflates the Occupied Grids to take into account the robot size. In slope inspection, we aim to develop an assisted obstacle avoidance system that can avoid obstacles even in unscanned areas, so the Unknown Grids should also be inflated like Occupied Grids to take into account the robot size. Such inflation is known as Unknown Grid Cells Inflation (Sec. \ref{sec:unk_inf}). Infinite Points Ray Casting (Sec. \ref{sec:inf_point}) tackles the issue of no LiDAR returned points when facing the sky. Incremental Frontiers Update (Sec. \ref{sec:map_frontier}) efficiently updates frontier information based on the latest sensor data. This process replaces a large number of Unknown grids with a small number of frontier grids, thereby reducing the computation time required for safe flight corridor (SFC) generation in path planning (Sec. \ref{sec:sfc}).

\subsubsection{Unknown Grid Cells Inflation}
\label{sec:unk_inf}

In motion planning, UAVs are often treated as point masses, and the occupied inflation radius $r_{occ}$ (\ie, user-defined obstacle avoidance distance) is employed to expand the obstacles in the inflated map. This inflation enables the UAV to avoid known obstacles during planning. However, existing methods often make the simplistic assumption that unknown areas are traversable. In reality, unknown areas may contain obstacles, presenting potential safety risks. To ensure a higher level of safety, it is necessary to not fly in unknown regions and further expand Unknown grids in the probability map by a radius $r_{unk}$ to consider the UAV size and safety clearance. Due to the Unknown Grid Cells Inflation, a grid cell state in the inflated map is redefined as Occupied Inflation, Unknown Inflation, and No Inflation.

\begin{table}[h]
    \centering
    \caption{Definitions of Grid Cell States in the Inflated Map}
    \label{tab:inf_grid}
    \begin{tabular}{|c|c|}
        \hline
        Grid Cell State & Definition \\
        \hline
        Occupied Inflation & $N_{occ} > 0$ \\
        \hline
        Unknown Inflation & $N_{occ} = 0$ and $N_{unk} > 0$ \\
        \hline
        No Inflation & $N_{occ} = 0$ and $N_{unk} = 0$ \\
        \hline
    \end{tabular}
\end{table}

\begin{figure}[h]
    \centering
    \includegraphics[width=\textwidth]{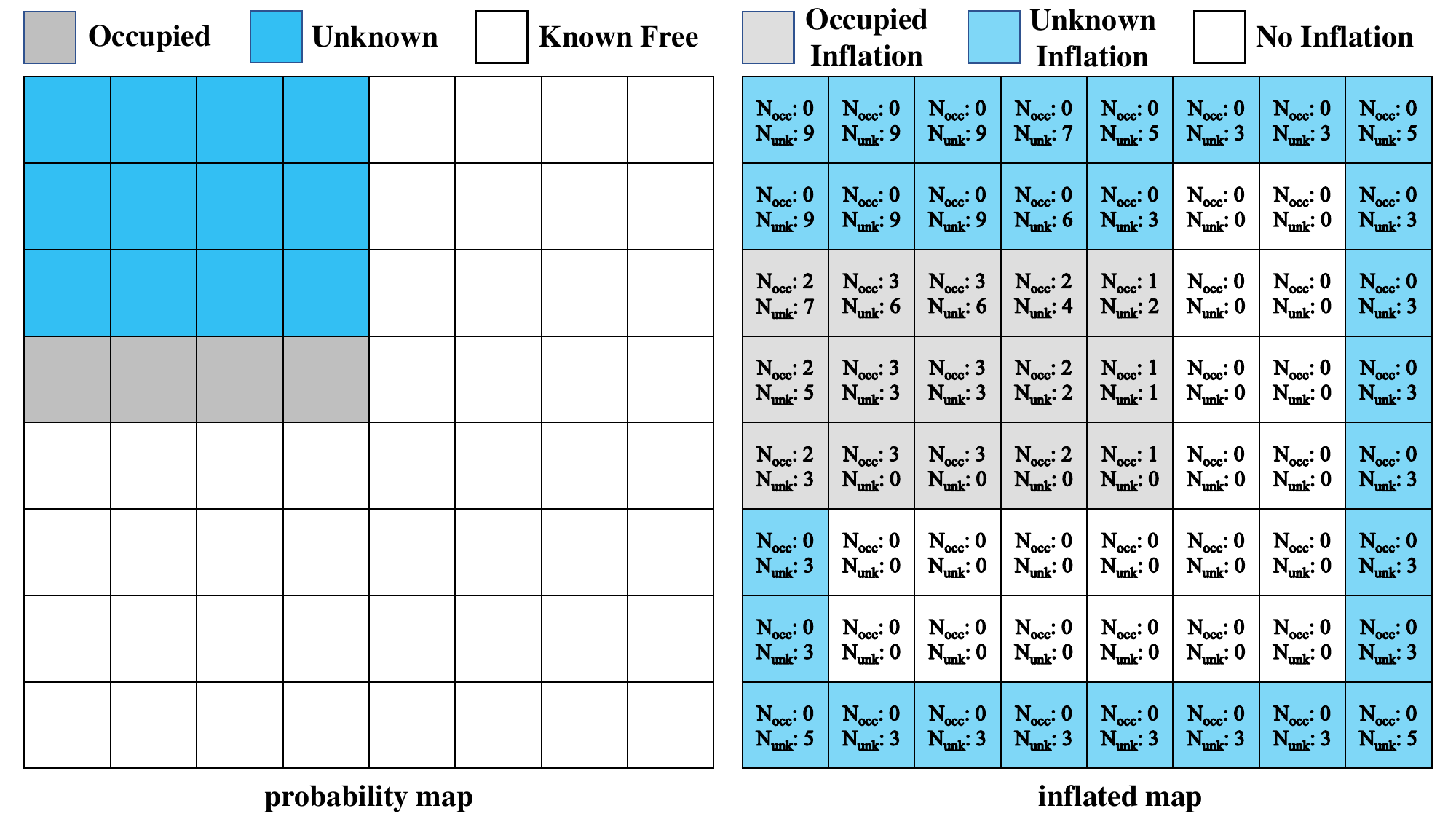}
    \caption{The Unknown Grid Cells Inflation in the 2D case. In this case, both the resolution of the probabilistic map and the inflated map are set to \SI{0.1}{m}, and the inflation radius, $r_{occ}$ and $r_{unk}$, are both set to \SI{0.2}{m}. As a result, the quantities of $\mathbf{I}_{occ}$ and $\mathbf{I}_{unk}$ are equal, with both being 9.}
    \label{fig:unk_inf}
\end{figure}

The method of Unknown Grid Cells Inflation is similar to the incremental update mechanism used in the original ROG-Map, which is achieved by maintaining a counter for each grid cell. Therefore, each grid cell in the inflated map maintains two non-negative counters, namely $N_{occ}$ and $N_{unk}$, which define the grid cell states as shown in Table \ref{tab:inf_grid}. Considering the inflation efficiency, it is necessary to determine the set of inflated grid cells for each Occupied or Unknown Grid Cell in advance. Such sets, denoted as $\mathbf{I}_{occ}$ and $\mathbf{I}_{unk}$, are expressed as grid cell coordinates offsets relative to the input Occupied or Unknown Grid Cell, and are computed offline from the respective inflation radius $r_{unk}$ and $r_{occ}$. Note that the unknown inflation radius $r_{unk}$ may differ from the occupied inflation radius $r_{occ}$ to give more flexibility. These offset coordinates are then added to the coordinates of the input grid cell to determine this grid cell's inflated occupied and unknown neighbors. During the initialization of the inflated map, all grid cells of the probability map are Unknown, therefore, $N_{occ}$ is set to 0, and $N_{unk}$ is set to the size of $\mathbf{I}_{unk}$ (including the grid cells at the map boundary since spaces outside the map boundary are also Unknown). Fig. \ref{fig:unk_inf} illustrates a simplified 2D process of updating from a probability map to an inflated map. Once occurring a change from an else state to Occupied in the probability map, the corresponding grid cell in the inflated map, along with its inflated neighboring grid cells defined by $\mathbf{I}_{occ}$, increments $N_{occ}$ by 1. Conversely, if the state changes from Occupied to else, $N_{occ}$ is decremented by 1. Similarly, when the state changes from an else state to Unknown, the corresponding grid cell in the inflated map, along with its inflated neighboring grid cells defined by $\mathbf{I}_{unk}$, increments $N_{unk}$ by 1. Conversely, if the state changes from Unknown to else, $N_{unk}$ is decremented by 1. By maintaining these two counters, we can effectively inflate the Unknown grid cells and update the inflated map in real-time based on changes of the grid cell state in the probability map.

\subsubsection{Infinite Points Ray Casting}
\label{sec:inf_point}

As an active sensor, LiDAR perceives the environment by emitting laser beams and receiving their return pulses. These laser beams that provide measurement are known as valid measurement points. However, certain situations can result in invalid measurement points. Firstly, when the LiDAR faces the sky, it is unable to measure distances due to the absence of returns. These LiDAR beams are referred to as infinite points. Secondly, when LiDAR scans nearby objects, the returned pulses are lumped into the pulses reflected by the LiDAR internal parts (\eg, prisms, glasses), causing the pulse return due to nearby objects to be indistinguishable from that due to the LiDAR internal parts. Points causing such a phenomenon are defined as nearby blind points. In the Livox Mid-360 LiDAR \cite{mid360} (and also many other LiDARs), the two cases are not distinguished, both leading to a point at LiDAR origin (\ie, invalid measurements). The inability to distinguish infinite points can cause a large number of grid cells lying in the direction of the sky or far buildings having their grid cell state not updated and remain Unknown. To update these grid cells to a Known Free state, we need to distinguish the infinite points from nearby blind points and fully utilize them for ray casting.

\begin{figure}[h]
    \centering
    \includegraphics[width=\textwidth]{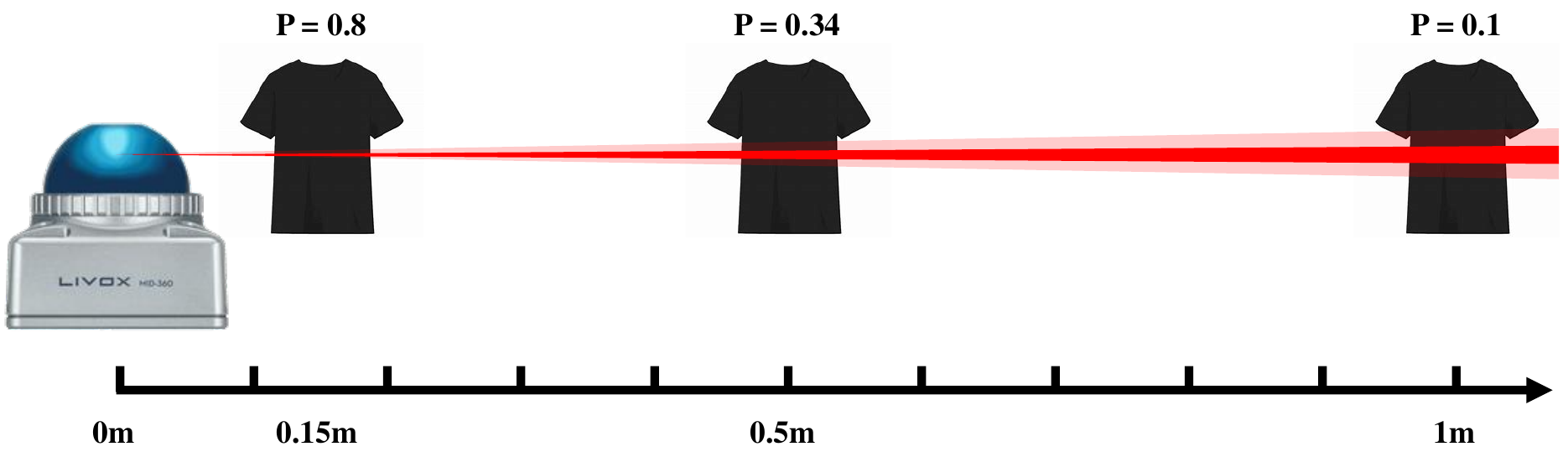}
    \caption{Characteristics of Livox Mid-360 LiDAR Scanning on Clothing. As the distance increases, the ratio of invalid measurement points to the total measurement points decreases. When the distance exceeds \SI{1}{m}, this ratio approaches zero, indicating that nearly all measurement points are considered valid.}
    \label{fig:inf_point}
\end{figure}

We analyze the characteristics of nearby blind points when scanning objects with the Livox Mid-360 LiDAR through an experiment. In the experiment, we scan a cloth at different distances within LiDAR's fixed field of view (FOV) and record the ratio of invalid measurement points to the total number of measurements. As shown in Fig. \ref{fig:inf_point}, when the cloth is located within a distance of \SI{1}{m}, the ratio of invalid measurements gradually decreases as the distance increases. Beyond \SI{1}{m}, the proportion stabilizes and approaches zero. These results indicate that the nearby blind points caused by objects within \SI{1}{m} account for approximately \SI{10}{\%} to \SI{80}{\%} invalid measurements, rendering the infinite points indistinguishable in such cases. Only when there are no close-proximate objects (\eg, within \SI{1}{m}), the invalid measurements can be considered as infinite points.

Based on the characteristics of nearby blind points analyzed above, we can extract infinite points by verifying the presence of close-proximate Occupied grid cells in the probability map. Specifically, when receiving LiDAR data, we extract the valid measurement points and invalid measurement points. For the valid measurement points, we perform ray casting to update the probabilities of the grid cells corresponding to obstacles. Next, for each invalid measurement point, we check if there are occupied grid cells within a distance of \SI{1}{m} along its ray direction. If no occupied grid cells are found within the 1-meter ray, we identify this invalid measurement point as an infinite point. Subsequently, within the local map region, we use these identified infinite points for ray casting to update the probability map. This process, known as Infinite Point Ray Casting, effectively extracts infinite points from the invalid measurement points, enabling the updating of probabilities for grid cells in the direction of the sky.

\subsubsection{Incremental Frontiers Update}
\label{sec:map_frontier}

\begin{figure}[h]
    \centering
    \includegraphics[width=\textwidth]{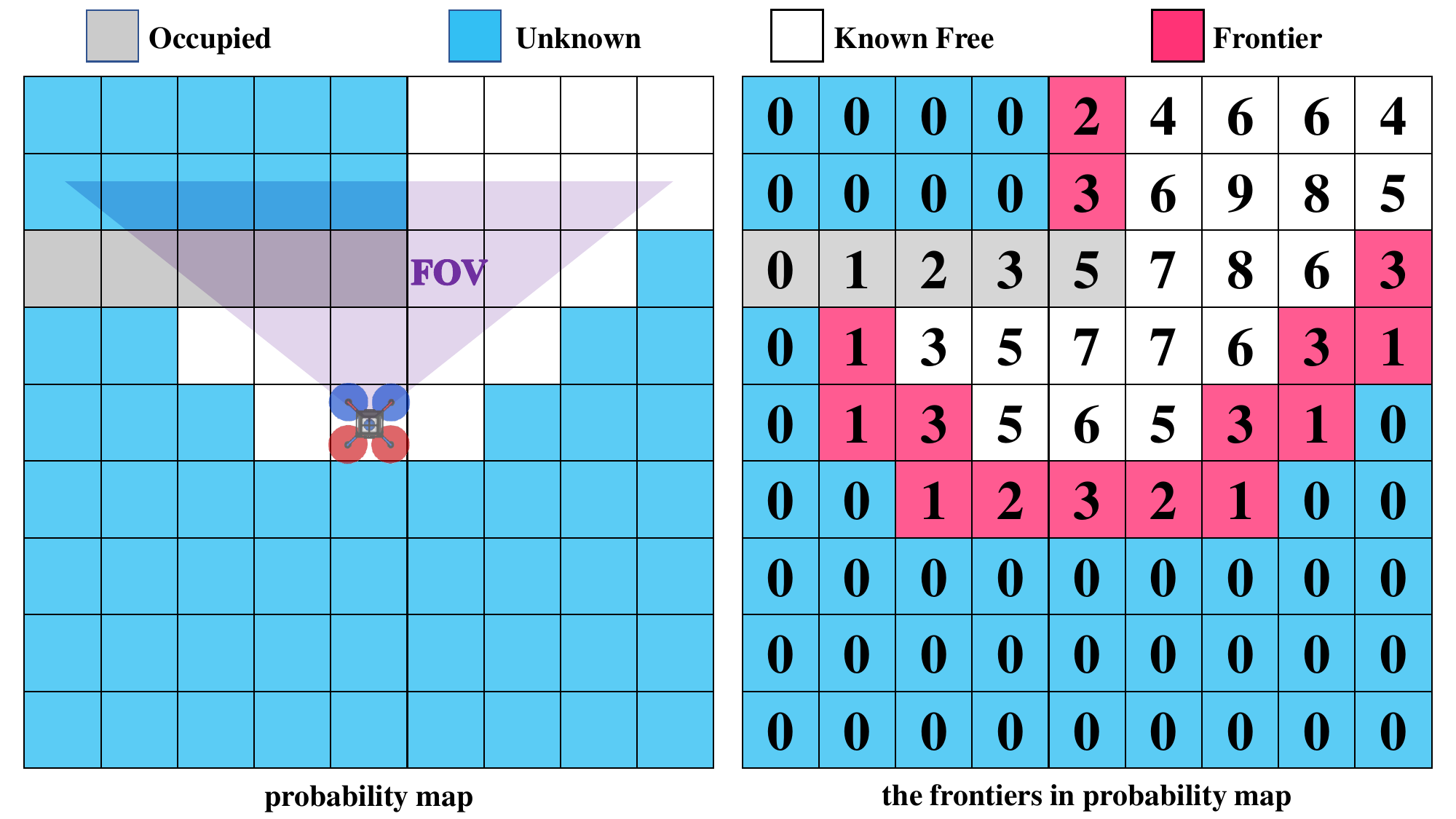}
    \caption{The Incremental Frontiers Update in the 2D case.}
    \label{fig:frontier}
\end{figure}

Due to the inability of the LiDAR sensor to scan the internal regions of obstacles through their surfaces and the limited LiDAR field of view (FOV), a large number of grid cells in the probabilistic map will be marked as Unknown. However, when generating the safe flight corridor (SFC) in the subsequent planning and control, a large number of unknown grids can considerably increase the computation time. Leveraging frontiers can effectively reduce the computation time required for SFC generation, as frontiers represent the boundary of the unknown region and are typically much fewer in quantity than Unknown grid cells. In the work \cite{yamauchi1997frontier}, frontiers are defined as Known Free grid cells adjacent to Unknown grid cells. We slightly modify this definition and define frontiers as Unknown grid cells adjacent to Known Free grid cells. Considering that SFC is generated in the probability map, we need to label the frontiers in the probability map. Specifically, we introduce a counter $N_f$ for each grid cell, indicating the number of {Known Free} grid cells among itself and its 26 neighbor grids. Fig. \ref{fig:frontier} illustrates a simplified 2D Incremental Frontiers Update process. During the initialization of the probabilistic map, all grid cells are Unknown (including areas outside the map range), therefore, $N_f$ is set to zero. When a grid cell state changes from Known Free to an else state, both the grid cell and its 26 neighboring grids decrement $N_f$ by 1. Conversely, when the grid cell state changes from an else state to Known Free, both the grid cell and its 26 neighboring grid cells increment $N_f$ by 1. If an Unknown grid cell has its $N_f$ less than 27 but greater than zero, this grid cell is classified as a frontier. Incremental Frontiers Update allows us to replace a large number of Unknown grids with a small number of frontier grids, thereby reducing the computation time required for SFC generation.

\subsection{Planning and Control}
\label{sec:ipc}

There are two primary modes of UAV-based inspection: fully autonomous inspection and human-in-the-loop inspection. Fully autonomous inspection relies on predefined flight rules and detection targets, making it suitable for inspection tasks with prior maps or knowledge of the detection targets. On the other hand, human-in-the-loop inspection offers greater flexibility and adaptability. In this mode, the UAV interprets the pilot's joystick commands as a local goal and reaches this goal while autonomously avoiding obstacles on the way. It is particularly well-suited for inspection tasks in unknown and complex environments, where no prior map is available or the inspection targets are decided impromptu during the flights. Considering the lack of prior maps or accurate coordinates of the targets in slope inspection, human-in-the-loop inspection is more suitable, and the pilot can command the UAV's local goal according to the intended inspection targets. Within the human-in-the-loop inspection mode, the assisted obstacle avoidance function in the planning and control module plays a critical role. The planning and control module generates actual control actuation for the UAV, such as throttle and angular velocity references, based on odometry, probability map, inflated map, and the pilot's joystick commands input, enabling safe flight in dense vegetation scenarios.

However, safe flight in dense vegetation poses several challenges for the planning and control modules: \textbf{1)} Avoiding thin objects effectively: The limited angular resolution of the LiDAR sensor results in a reduced sensing range for thin objects, such as branches and vines. To ensure safe flight, the quadrotor must react promptly within this shorter sensing range to avoid thin objects. \textbf{2)} Natural wind disturbances: In outdoor environments, the quadrotor may be exposed to wind disturbances, which require the quadrotor to possess robust disturbance rejection capabilities to maintain stability during flight. \textbf{3)} Chaotic joystick signals: Joystick signals from the pilot's remote controller can exhibit erratic behavior, leading to high-frequency variations in the reference position. The quadrotor must respond rapidly to these signals to accurately execute pilot intention. \textbf{4)} Limited onboard computational resources: Due to compact size and weight restrictions, the available onboard computational resources on the quadrotor are limited. The planning and control module must be efficient and capable of generating control actions within milliseconds.

Most quadrotor navigation approaches \cite{liu2017planning,zhou2019robust,zhang2020falco,zhou2020ego,tordesillas2021faster,ren2022bubble,kim2023autonomous,mellinger2011minimum,ren2023online} typically employ a planning and control separation framework. In this framework, The planner generates high-order smooth trajectories that adhere to the dynamical constraints within a safe space, while the controller produces control actuation to track the trajectory reference. However, this multi-stage pipeline results in increased system latency, and the planner does not consider disturbances, resulting in an inability to respond promptly to disturbances (\eg, wind gusts), subsequently affecting the safe flight of the quadrotor. Additionally, the erratic joystick commands from the pilot lead to frequent changes in local goals, causing a mismatch between the high-order trajectories generated by the planner and the quadrotor's expected actions. As a result, the quadrotor tends to exhibit overly conservative flight behavior, reducing the responsiveness to the pilot's commands.

Our previous work IPC \cite{liu2023integrated} presents an integrated planning and control framework that effectively tackles the aforementioned challenges. IPC enables real-time computation of control actions and trajectories for quadrotor at a frequency of \SI{100}{Hz}, allowing for rapid response to dynamic obstacles. Moreover, by integrating planning and control within the Model Predictive Control (MPC) problem, the IPC backend enhances its ability to suppress external disturbances. Additionally, IPC does not impose high-order trajectory constraints, allowing quadrotor to exhibit more aggressive flight behaviors that better align with the pilot's commands.

\begin{figure}[h]
    \centering
    \includegraphics[width=\textwidth]{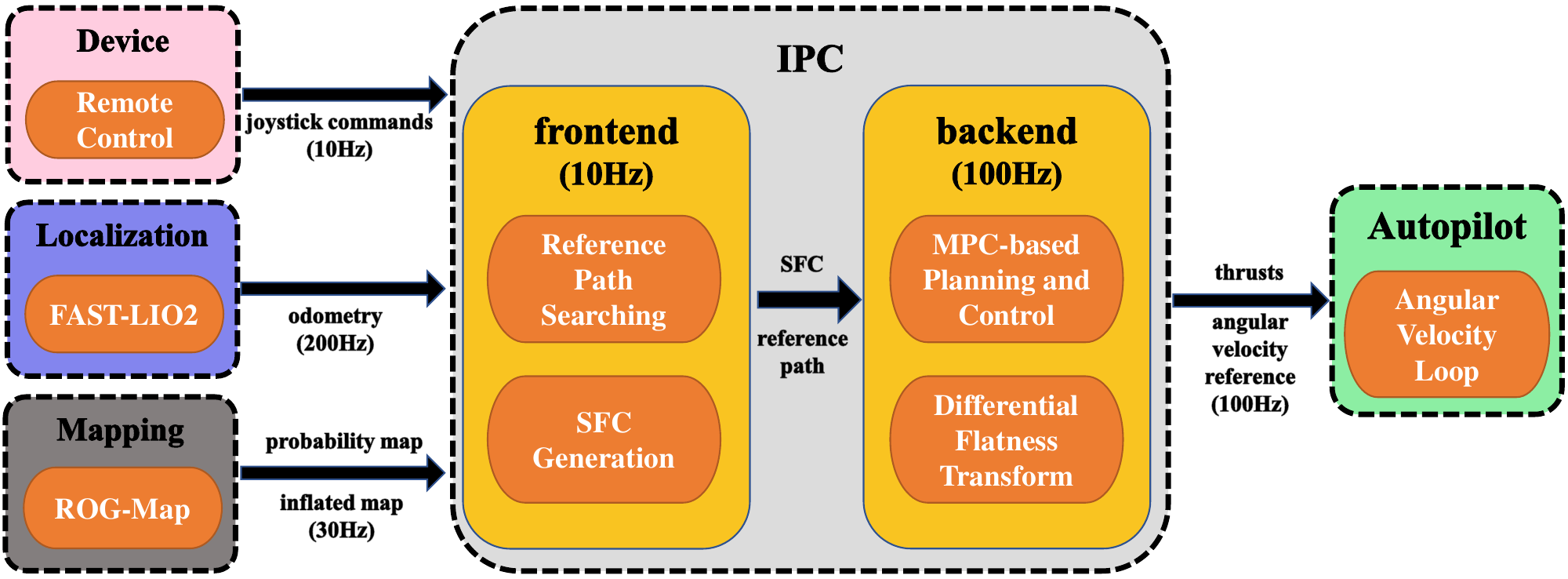}
    \caption{The IPC structure. IPC is an integrated planning and control framework consisting of the frontend and backend. By inputting the joystick commands, odometry, probability map and inflated map, the IPC outputs the angular velocity reference and throttle commands that are further tracked by the quadrotor's onboard autopilot.}
    \label{fig:ipc_struct}
\end{figure}

The original IPC considers autonomous flights with a given goal position that is constant during the flights, lacking the incorporation of the pilot's joystick commands input. Moreover, IPC treats the areas not scanned by sensors (\ie, the Unknown grid cells of the probability map) as obstacle-free, lacking a rigorous safety guarantee. Additionally, we do not incorporate any design for assisted flight based on pilot joystick input. To achieve assisted obstacle avoidance flight in dense vegetation, we redesign IPC's frontend in this work. This redesign aims to incorporate pilot joystick input and improve safety guarantees. The updated IPC framework is depicted in Fig. \ref{fig:ipc_struct}.

\subsubsection{Reference Path Searching}
\label{sec:ref_path}

\begin{algorithm}[t]
    \SetAlgoLined
    \footnotesize
    \caption{Reference Path Searching}
    \label{alg:path}
    \KwIn{inflated map $\Theta$, quadrotor's current position ${\mathbf{p}}_{odom}$, local goal ${\mathbf{p}}_{g}$, maximum search radius $\beta$.}
    \KwOut{reference path $\mathbf{P}$, No Inflation reference path $\mathbf{P}_{no\_inf}$.}  
    \BlankLine
    ${\mathbf{p}}_{s} \leftarrow {\mathbf{p}}_{odom}$\; \label{alg:path:ps}
    $\mathbf{P}_{inf}.\textnormal{clear}()$\; \label{alg:path:p_clear}
    \If{$\mathtt{IsInflationInMap}({\mathbf{p}}_{odom}, \Theta)$}  { 
        ${\mathbf{p}}_{s}, \mathbf{P}_{inf} \leftarrow \mathtt{BreadthFirstSearch}({\mathbf{p}}_{odom})$\; \label{alg:path:bfs}
    } \label{alg:path:new_ps}
    \If{$!\mathtt{IsInflationInMap}({\mathbf{p}}_{g}, \Theta)$} { \label{alg:path:pg_joy_s}
        ${\mathbf{p}}_{g}, \mathbf{P}_{no\_inf} \leftarrow \mathtt{FindFarestGrid}({\mathbf{p}}_{s}, {\mathbf{p}}_{g})$\; \label{alg:path:FindFarestGrid}
    }
    \Else {
        ${\mathbf{p}}_{gn}, \mathbf{P}_{f} \leftarrow \mathtt{BreadthFirstSearch}({\mathbf{p}}_{g})$\; \label{alg:path:bfs_n}
        ${\mathbf{p}}_{g}, \mathbf{P}_{no\_inf} \leftarrow \mathtt{FindFarestGrid}({\mathbf{p}}_{s}, {\mathbf{p}}_{gn})$\; \label{alg:path:FindFarestGrid_n}
    } \label{alg:path:pg_joy_e}
    $\mathbf{P} \leftarrow \{ \mathbf{P}_{inf}, \mathbf{P}_{no\_inf} \}$\; \label{alg:path:p}
\end{algorithm}

\begin{figure}[h]
    \centering
    \includegraphics[width=0.95\textwidth]{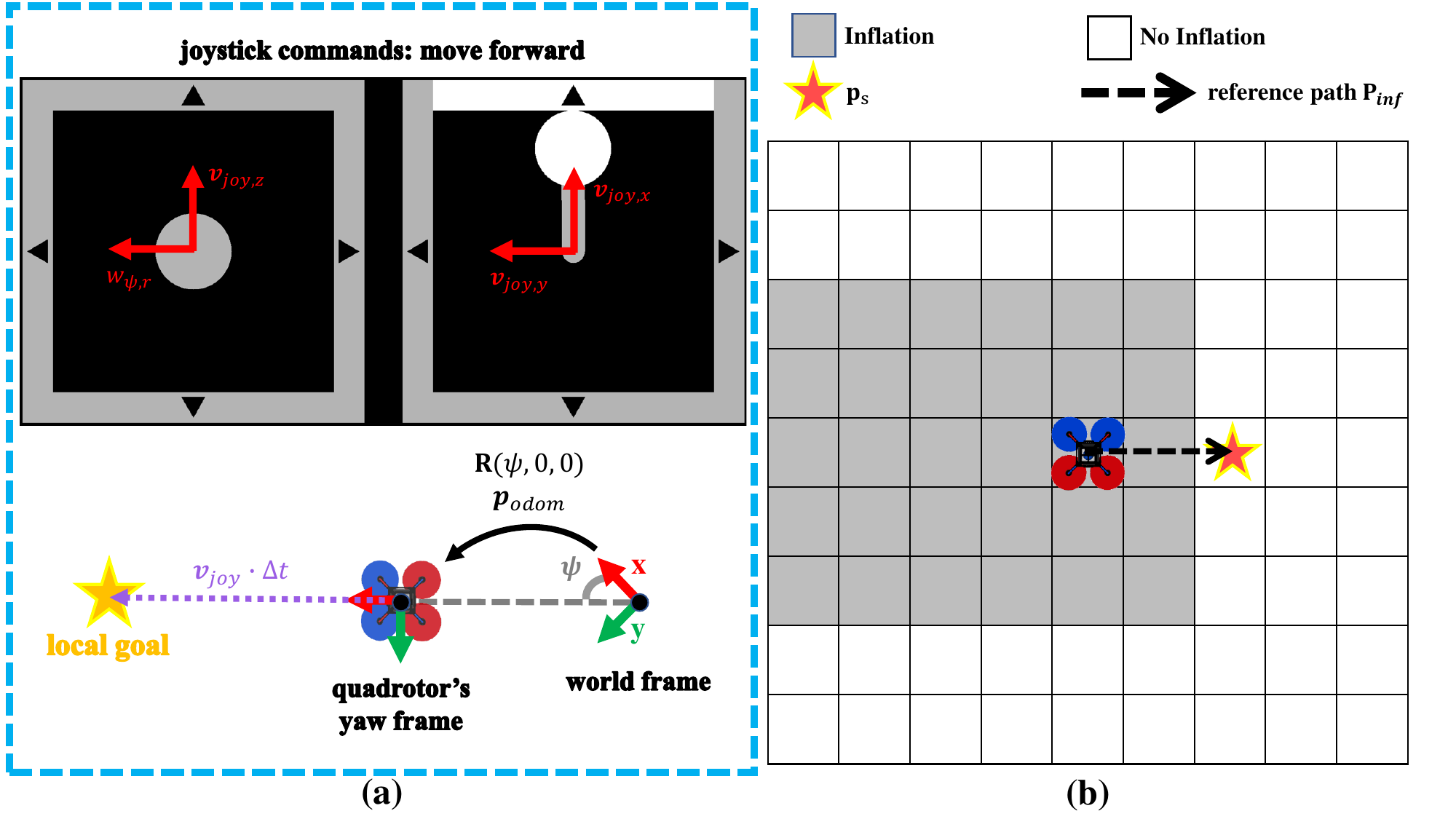}
    \caption{(a) Mapping the joystick commands to reference velocity in the quadrotor's yaw frame ${\mathbf{v}}_{joy}$ and reference yaw angular velocity $w_{\psi,r}$. Then the local goal ${\mathbf{p}}_{g}$ and the quadrotor's yaw reference $\psi_{r}$ is identified. (b) The case of the quadrotor's current position being in an Inflation (Occupied Inflation or Unknown Inflation) grid cell, a reference path $\mathbf{P}_{inf}$ is first searched to leave the inflation area.}
    \label{fig:ref_path1}
\end{figure}

The first step of reference path searching is to identify a local goal ${\mathbf{p}}_{g}$ from the pilot's joystick commands. As shown in Fig. \ref{fig:ref_path1}(a), the joystick commands consists of velocity commands ${\mathbf{v}}_{joy} \in \mathbb{R}^3$ and yaw rate command $w_{\psi,r} \in \mathbb{R}$. We consider the velocity command ${\mathbf{v}}_{joy}$ as specified in the quadrotor's yaw frame (a frame with only a rotation in the yaw direction to the world frame, while pitch and roll are set to zero). Let $\mathbf{R}(\psi, 0, 0)$ be the rotation of the yaw frame with respect to the world frame, where $\psi$ is the quadrotor's yaw angle, the local goal in the world frame can be computed as:
\begin{equation}
    {\mathbf{p}}_{g} = \mathbf{R}(\psi, 0, 0) \cdot {\mathbf{v}}_{joy} \cdot \triangle{t} + {\mathbf{p}}_{odom}
    \label{eq:joy}
\end{equation}
where ${\mathbf{p}}_{odom}$ is the quadrotor's current position, $\triangle{t}$ is the joystick commands period (\eg, $\triangle{t}$ is \SI{0.1}{s} when the joystick commands frequency is \SI{10}{Hz}).

For the yaw rate command $w_{\psi,r}$, it is mapped to the desired quadrotor's yaw angle $\psi_{r}$ as:
\begin{equation}
    \psi_{r} = \psi + w_{\psi,r} \cdot \triangle{t}
    \label{eq:yaw}
\end{equation}
where $\psi$ denotes the current quadrotor's yaw angle.

After finding the local goal, Reference Path Searching aims to search for a feasible path from the quadrotor's current position ${\mathbf{p}}_{odom}$ to the local goal ${\mathbf{p}}_{g}$ in the inflated map $\Theta$ (Sec. \ref{sec:unk_inf}). The Reference Path Searching, running at \SI{10}{Hz}, is illustrated in Alg. \ref{alg:path}. The searched reference path $\mathbf{P}$ consists of two segments: 

1. \textbf{The reference path in the Inflation region $\mathbf{P}_{inf}$} (Lines \ref{alg:path:ps}-\ref{alg:path:new_ps}): When encountering dynamic objects or control errors, the quadrotor's current position ${\mathbf{p}}_{odom}$ may accidentally be in Inflation state within the inflated map $\Theta$, making it impossible to find a feasible path (\ie, a path in the No Inflation area). To fix this issue, it is necessary to search for a path that swiftly navigates the quadrotor to a position $\mathbf p_s$ in No Inflation regions. Initially, ${\mathbf{p}}_{s}$ is set to ${\mathbf{p}}_{odom}$ (Line \ref{alg:path:ps}) and $\mathbf{P}_{inf}$ remains empty (Line \ref{alg:path:p_clear}). If ${\mathbf{p}}_{odom}$ is in Inflation state in $\Theta$ (Fig. \ref{fig:ref_path1}(b)), a breadth-first search is conducted until encounter the first No Inflation grid cell or time out (Line \ref{alg:path:bfs}). The first encountered No Inflation grid cell is the ${\mathbf{p}}_{s}$, and the path to it is the $\mathbf P_{inf}$.

\begin{figure}[h]
    \centering
    \includegraphics[width=0.95\textwidth]{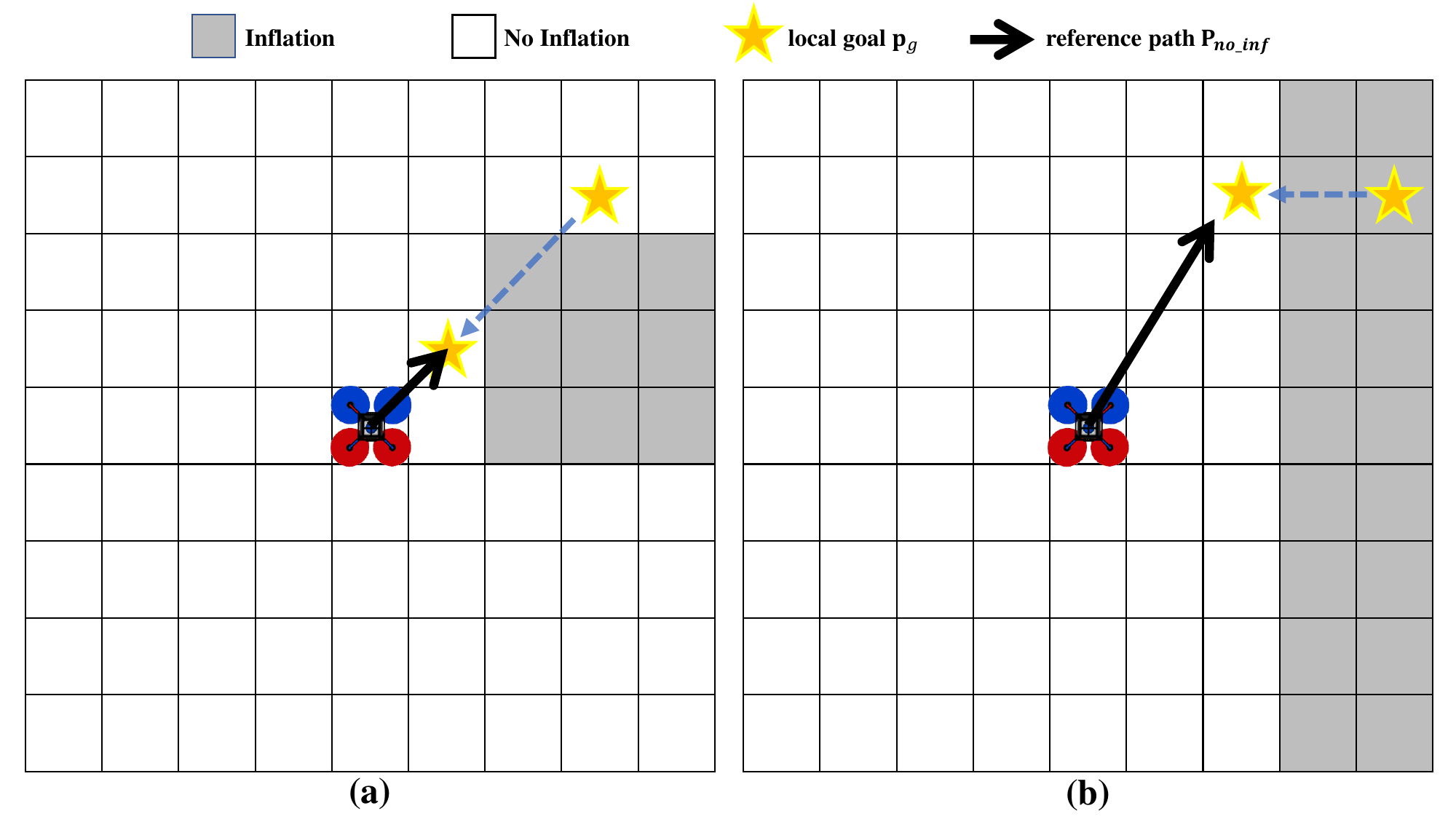}
    \caption{{Two cases of reference path $\mathbf{P}_{no\_inf}$ searching. (a) If the $\mathbf{p}_{g}$ is in No Inflation state but occluded by Inflation grid cells, the farthest visible No Inflation grid cell will be identified as the new $\mathbf{p}_{g}$. (b) If the $\mathbf{p}_{g}$ is in Inflation state, a nearby No Inflation grid cell will be identified as the new $\mathbf{p}_{g}$.}}
    \label{fig:ref_path2}
\end{figure}

2. \textbf{The reference path in the No Inflation region $\mathbf{P}_{no\_inf}$} (Lines \ref{alg:path:pg_joy_s}-\ref{alg:path:pg_joy_e}): If ${\mathbf{p}}_{g}$ is in No Inflation state in $\Theta$, but there may be Inflation grid cells on the line segment between ${\mathbf{p}}_{s}$ and ${\mathbf{p}}_{g}$ (Fig. \ref{fig:ref_path2}(a)), the farthest No Inflation grid cell on this line segment is used as the final local goal ${\mathbf{p}}_{g}$, and the straight line from ${\mathbf{p}}_{s}$ to ${\mathbf{p}}_{g}$ is returned as the path $\mathbf P_{no\_inf}$ (Line \ref{alg:path:FindFarestGrid}). If ${\mathbf{p}}_{g}$ is in Inflation state in $\Theta$ (Fig. \ref{fig:ref_path2}(b)), another breadth-first search similar to the starting position is conducted to obtain a feasible local goal ${\mathbf{p}}_{gn}$ (Line \ref{alg:path:bfs_n}). Then, the feasible local goal ${\mathbf{p}}_{gn}$ is further modified as in the first case to obtain the final local goal ${\mathbf{p}}_{g}$, which is the farthest No Inflation point from ${\mathbf{p}}_{s}$ to ${\mathbf{p}}_{gn}$, and the path $\mathbf P_{no\_inf}$ (Line \ref{alg:path:FindFarestGrid_n}). Finally, the complete reference path $\mathbf{P}$ is obtained as the union of the two path segments $\mathbf P_{inf}$ and $\mathbf P_{no\_inf}$ (Line \ref{alg:path:p}).

It is worth noting that in the inflated map (Sec. \ref{sec:unk_inf}), we did not track or predict the dynamic objects' movements. Instead, dynamic objects are treated together with static obstacles and subjected to inflation operations. The lack of dynamic objects tracking and prediction are compensated by the high planning and control rate of our overall framework, which can avoid dynamic objects in a purely reactive manner. While a larger inflation radius can provide the quadrotor with more reaction time and distance to avoid dynamic objects, it significantly reduces the available space in the inflated map, consequently decreasing maneuverability in narrow areas. Therefore, different obstacle avoidance requirements can be met by adjusting the occupied inflation radius $r_{occ}$ (\ie, the user-defined obstacle avoidance distance, $d_0$) in the inflated map (Sec. \ref{sec:unk_inf}). For instance, setting $d_0$ to approach the quadrotor's radius allows for flight in narrow areas. Alternatively, setting $d_0$ to three times the quadrotor's radius can effectively evade dynamic objects.

\subsubsection{SFC Generation}
\label{sec:sfc}

\begin{figure}[h]
    \centering
    \includegraphics[width=\textwidth]{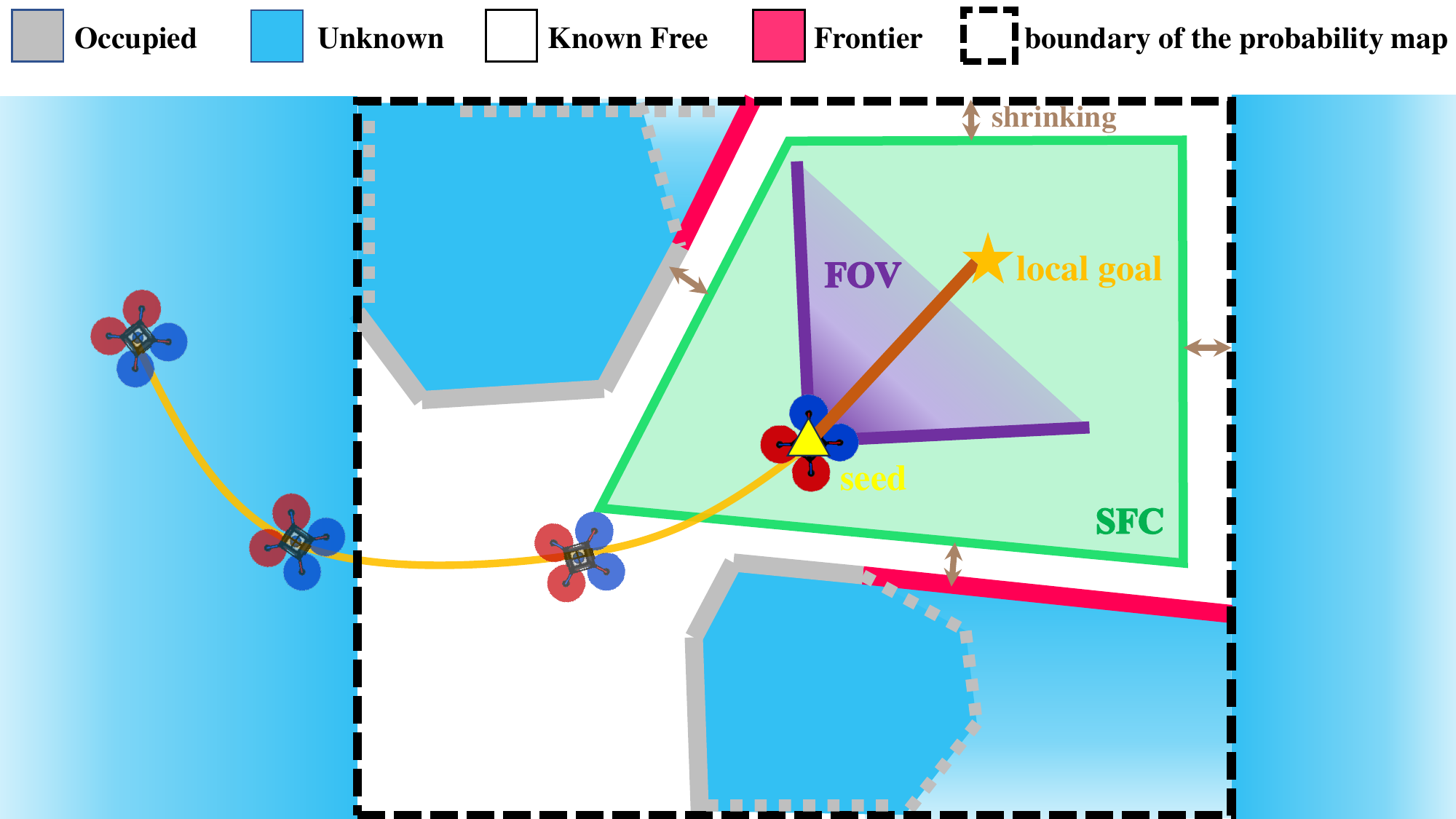}
    \caption{The SFC generation considering the unknown areas: The grid cells along the pink lines represent Frontiers, while the grid cells along the gray lines represent Occupied. The brown line represents the reference path $\mathbf{P}$, and the yellow triangle denotes the seed for SFC generation. The orange curve represents the quadrotor's flight trajectory, while the green polygon represents the generated SFC.}
    \label{fig:sfc}
\end{figure}

After the Reference Path Searching (Sec. \ref{sec:ref_path}), we directly adopt the method proposed in \cite{liu2017planning} to generate safe flight corridor (SFC) along the reference path. Since no SFC can be generated along the reference path $\mathbf P_{inf}$ in the Inflation area, we only consider generating SFC along the No Inflation reference path $\mathbf P_{no\_inf}$, which resides in the Known Free space of the probability map (Sec. \ref{sec:map}). Considering that the quadrotor's flight speed in narrow spaces is low, we generate only one convex polyhedron as the SFC to reduce the computation load. We input the grid cells that are identified as Occupied or Frontiers, instead of Occupied or Unknown for improved efficiency, along with the seed, which is the point on the path $\mathbf{P}_{no\_inf}$ that is closest to the quadrotor's current position, to the polyhedron generation method \cite{liu2017planning}. The obtained polyhedron will intersect with the local map boundary (since the spaces enclosed by Occupied and Frontiers are Known Free when they are within the local map, see Fig. \ref{fig:frontier}) and shrink by the quadrotor's size. A simplified 2D scenario for SFC generation is illustrated in Fig. \ref{fig:sfc}. 

\subsubsection{Backend}
\label{sec:backend}

In the backend, we directly employ two parts from our previous work \cite{liu2023integrated}: Model Predictive Control(MPC)-based Planning and Control and Differential Flatness Transform. The goal of the MPC is to guide the quadrotor along the reference path $\mathbf{P}$ (Sec. \ref{sec:ref_path}) at preset reference speed $v_r$, while keeping the quadrotor in the free space represented by SFC (Sec. \ref{sec:sfc}) and satisfying necessary constraints. The Differential Flatness Transform is responsible for converting the MPC optimization variables into actual angular velocity references and thrusts. This transformation allows for directly control of the quadrotor's rotor speed through a lower-level angular velocity controller, enabling the quadrotor's complete motion of the free space.

The symbols used in MPC are defined in Table \ref{table:define}. In the system model presented in MPC, the system state of the quadrotor is denoted by {$\mathbf{x} = [\mathbf{p}, \mathbf{v}, \mathbf{a}]^T$} and the system input is represented by {$\mathbf{u}=\mathbf{j}$}.

\begin{table}[h]
    \centering
    \caption{Nomenclature}
    \begin{tabular}{c c}
        \hline
        $\mathbf{p}$ & \makecell[l]{position vector $p_x,p_y,p_z$ in the world frame} \\
        $\mathbf{v}$ & \makecell[l]{velocity vector $v_x,v_y,v_z$ in the world frame} \\
        $\mathbf{a}$ & \makecell[l]{acceleration vector $a_x,a_y,a_z$ in the world frame} \\
        $\mathbf{j}$ & \makecell[l]{jerk vector $j_x,j_y,j_z$ in the world frame} \\
        $N$ & \makecell[l]{horizon length in the MPC} \\
        $\triangle{t}$ & \makecell[l]{time step of the MPC} \\
        \hline
    \end{tabular}
    \label{table:define}
\end{table}

In order for the MPC to follow the reference path $\mathbf{P}$, we sample $N$, the horizon length of the MPC, reference positions $\mathbf p_{ref,n}, n = 1, 2, ..., N$, on the reference path $\mathbf{P}$. The first reference position $\mathbf p_{ref,1}$ is the position on $\mathbf{P}$ that is closest to the quadrotor's current position $\mathbf{p}_{odom}$, which is obtained by odometry (Sec. \ref{sec:lio}). Starting from $\mathbf{p}_{ref,1}$, we sample waypoints at intervals of $v_r*\Delta t$ (where $\Delta t$ represents the model discretization time in MPC) on the reference path, each sampled waypoint is added to the set of reference positions $\mathbf p_{ref,n}$. We continue this process until we have sampled $N$ waypoints, the sampled waypoint reaches the end of the reference path, or the sampled waypoint falls out of the SFC. In the later two cases, the remaining reference positions will be set to the last valid sampled waypoint, so the total length of the reference positions is also $N$.

With the reference positions $\mathbf p_{ref,n}, n = 1, 2, ..., N$ obtained above, our MPC is formulated as:
\begin{subequations}
    \begin{align}
        \begin{split}
            \underset{\mathbf{u}_k}{{\min}} \quad &\sum_{n=1}^N({\left\|(\mathbf{p}_{ref,n}-\mathbf{p}_n)\right\|}_{\mathbf{R}_p}^2 + {\left\|\mathbf{u}_{n-1}\right\|}_{\mathbf{R}_u}^2 ) \\ 
            + &{\left\|\mathbf{v}_N\right\|}_{\mathbf{R}_{v,N}}^2 + {\left\|\mathbf{a}_N\right\|}_{\mathbf{R}_{a,N}}^2 + \sum_{n=0}^{N-2}{\left\|\mathbf{u}_{n+1}-\mathbf{u}_{n}\right\|}_{\mathbf{R}_c}^2 \label{eq:cost}
        \end{split} \\
        \mathbf{s.t.} \quad & \mathbf{x}_{n} = \mathbf f_d(\mathbf{x}_{n-1}, \mathbf{u}_{n-1}), \quad n = 1, 2, \cdots, N \label{eq:st_model} \\
        & \mathbf{x}_0 = [\mathbf{p}_{odom}, \mathbf{v}_{odom}, \mathbf{a}_{odom}]^T \label{eq:init} \\
        & |v_{i,n}| \leq |v_{i,max}|, i = x,y,z \label{eq:st1} \\
        & |a_{j,n}| \leq |a_{j,max}|, j = x,y \label{eq:st2} \\
        & a_{z,min} \leq a_{z,n} \leq a_{z,max} \label{eq:st3} \\
        & |j_{i,n}| \leq |j_{i,max}|, i = x,y,z \label{eq:st4} \\
        & \mathbf{C} \cdot \mathbf{p}_{n} - \mathbf{d} \leq 0 \label{eq:st_pos}
    \end{align}
    \label{eq:mpc}
\end{subequations}
where the cost function (\ref{eq:cost}) consists of ${\left\|\mathbf{p}_{ref,n}-\mathbf{p}_n\right\|}_{\mathbf{R}_p}^2$, the reference path following error, ${\left\|\mathbf{u}_{n-1}\right\|}_{\mathbf{R}_u}^2$, the control efforts, $\left\|\mathbf{u}_{n+1}-\mathbf{u}_{n}\right\|_{\mathbf R_c}^2$, the control variation, ${\left\|\mathbf{v}_N\right\|}_{\mathbf{R}_{v,N}}^2$, the terminal velocity, and ${\left\|\mathbf{a}_N\right\|}_{\mathbf{R}_{a,N}}^2$, the terminal acceleration. 

The constraints in the formulated MPC problem (\ref{eq:mpc}) consist of three. The first one is the model constraints (\ref{eq:st_model}) subject to initial state (\ref{eq:init}) estimated by an odometry. To reduce the MPC complexity, we adopt a third-order integrator for the quadrotor:
\begin{equation}
    \begin{aligned}
        \mathbf{p}_{n} &= \mathbf{p}_{n-1} + \triangle{t} \cdot \mathbf{v}_{n-1} + \frac{1}{2}\triangle{t}^2 \cdot \mathbf{a}_{n-1} + \frac{1}{6}\triangle{t}^3 \cdot \mathbf{j}_{n-1} \\
        \mathbf{v}_{n} &= \mathbf{v}_{n-1} + \triangle{t} \cdot \mathbf{a}_{n-1} + \frac{1}{2}\triangle{t}^2 \cdot \mathbf{j}_{n-1} \\
        \mathbf{a}_{n} &= \mathbf{a}_{n-1} + \triangle{t} \cdot \mathbf{j}_{n-1} \\
        \mathbf{x}_{n} &= [\mathbf{p}_n, \mathbf{v}_n, \mathbf{a}_n]^T, \quad \mathbf{u}_n = \mathbf{j}_n
    \end{aligned}
    \label{eq:sys_dis}
\end{equation}

The second constraints are the kinodynamic constraints (\ref{eq:st1}-\ref{eq:st4}), which ensure the quadrotor's dynamics are within feasible limits. The third constraints are the corridor constraints (\ref{eq:st_pos}), which ensure the quadrotor to remain within the safe flight corridor, which is represented as $\{ \mathbf p \in \mathbb{R}^3 | \mathbf C \cdot \mathbf p \preceq \mathbf d\}$, hence avoiding collision with both dynamic and static obstacles in the environments.

The optimization problem (\ref{eq:mpc}) involves a quadratic cost and linear constraints in terms of the optimization variables $\mathbf{U} = [\mathbf{u}_0, \mathbf{u}_1, ... , \mathbf{u}_{N-1}]^T$, which presents a standard quadratic programming (QP) problem. This QP problem is solved by OSQP-Eigen\footnote{\href{https://github.com/robotology/osqp-eigen}{https://github.com/robotology/osqp-eigen}}, a C++ library that depends on OSQP \cite{osqp} and Eigen3 \cite{eigenweb}. The resulting solution generates the optimal control actions and local trajectory according to the cost function.

After solving the MPC problem (\ref{eq:mpc}), the optimal control actions $\mathbf{j}$, defined in the world frame, cannot be directly applied to the quadrotor in the real world because it is not the commands to the quadrotor actuators (\ie, motors). Therefore, we utilize the differential flatness property \cite{mellinger2011minimum} of the quadrotor to transform the jerk $\mathbf{j}$ along with other states such as acceleration {$\mathbf{a}$} into angular velocity reference. The angular velocity reference is finally tracked by lower-level controllers implemented onboard the autopilot to produce the motor commands. 
\begin{subequations}
    \begin{align}
        & \mathbf{z}_B = \frac{\mathbf{t}}{\left\|\mathbf{t}\right\|}, \quad \mathbf{t} = [a_x, a_y, a_z+g]^T \\
        & \mathbf{x}_C = [\cos{\psi}, \sin{\psi}, 0]^T \\
        & \mathbf{y}_B = \frac{\mathbf{z}_B \times \mathbf{x}_C}{\left\| \mathbf{z}_B \times \mathbf{x}_C \right\|}, \quad \mathbf{x}_B = \mathbf{y}_B \times \mathbf{z}_B \\
        & \mathbf{h}_w = \frac{(\mathbf{j} - (\mathbf{z}_B \cdot \mathbf{j})\mathbf{z}_B)}{\left\|\mathbf{a}\right\|} \\
        & p_r = -\mathbf{h}_w \cdot \mathbf{y}_B, \quad q_r = \mathbf{h}_w \cdot \mathbf{x}_B \\
        & r_r = ({\psi}_r - \psi) \cdot \mathbf{z}_B \cdot (0,0,1)^T
    \end{align}
    \label{eq:flat}
\end{subequations}
where $g$ represents the gravitational acceleration, ${\psi}_r$ and $\psi$ are the reference and feedback of the quadrotor's yaw angle in the world frame, $(p_r,q_r,r_r)$ denote the pitch, roll and yaw angular velocity reference in the body frame. 

In addition, we also need to calculate the throttle $T_r$ of the quadrotor to control its motion along the Z-axis:
\begin{equation}
    T_r = C_T \cdot \left\|\mathbf{t}\right\|
    \label{eq:throttle}
\end{equation}
where $C_T$ is the throttle thrust coefficient that is calibrated beforehand.

\section{Experiments}
\label{sec:experiments}

To validate the applicability of our LiDAR-based quadrotor for slope inspection, we conduct a series of comprehensive tests and experiments. In Sec. \ref{sec:indoor_test}, the assisted obstacle avoidance flight function is evaluated in non-operational scenarios. Subsequently, our quadrotor is deployed to six actual slopes with dense vegetation in Hong Kong, as detailed in Sec. \ref{sec:field_test}. Throughout all the six field tests, our quadrotor successfully performs close-up photo inspections of flexible debris-resisting barriers while navigating safely in the complex environments. Additionally, in Sec. \ref{sec:benchmark}, we conduct comparative experiments in the field environments with DJI Mavic 3 to further showcase the assisted obstacle avoidance capabilities of our quadrotor.

\subsection{Functional Tests in Non-Operational Scenarios}
\label{sec:indoor_test}

\begin{figure}[h]
    \centering
    \includegraphics[width=\textwidth]{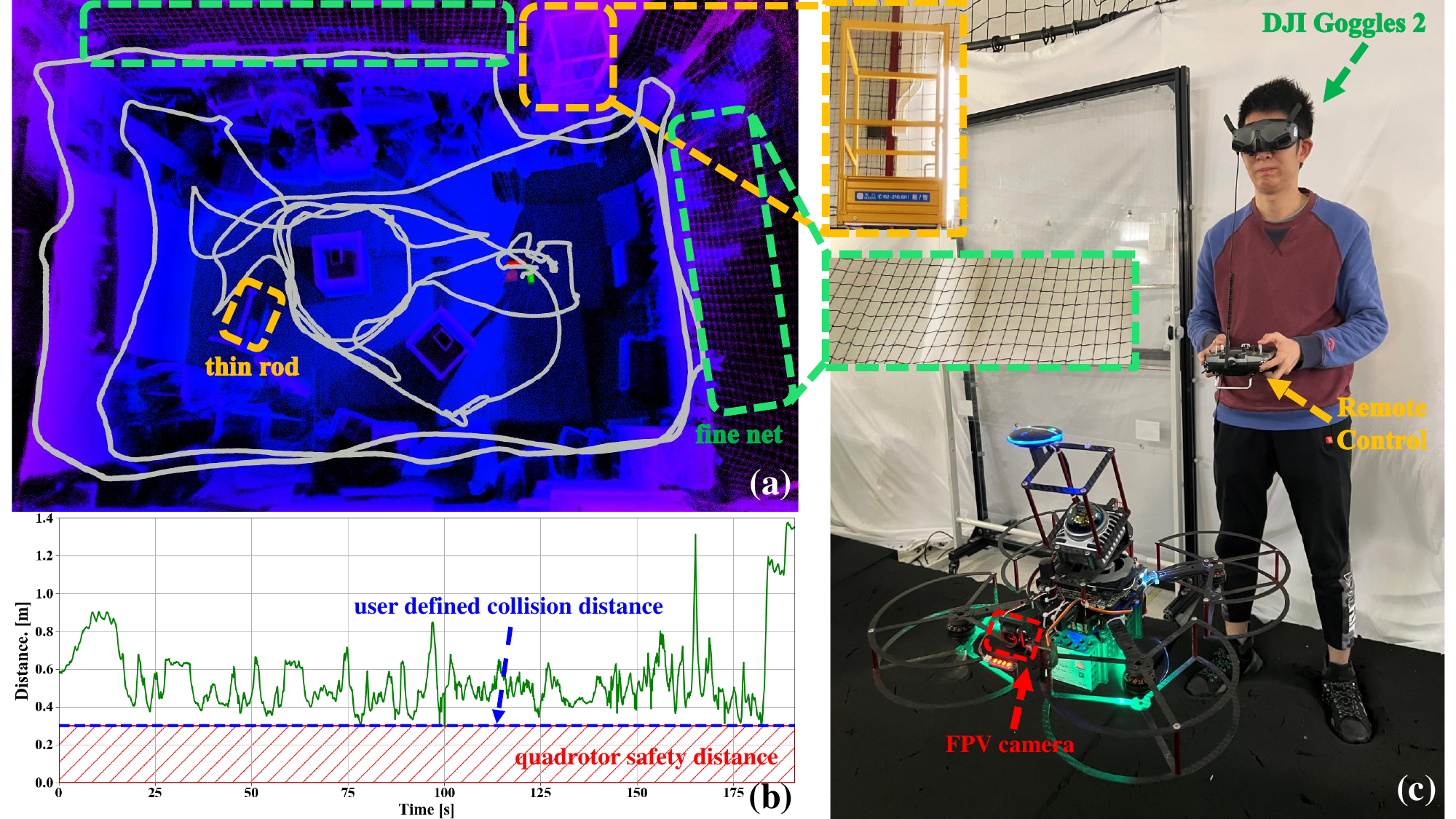}
    \caption{Assisted obstacle avoidance flight in a narrow environment. (a) The gray curve represents the quadrotor's flight trajectory, which is about \SI{99.46}{m} long. The green box represents the fine nets, and the orange box represents a scaffold with thin rods. (b) The distance between the quadrotor's center and the nearest obstacles during the flight. (c) The third-person view of the obstacle avoidance flight with the pilot wearing the DJI Goggles 2. The pilot is operating the quadrotor purely based on the FPV video seen in the Goggle.}
    \label{fig:indoor}
\end{figure}

We conduct functional tests of the quadrotor's assisted obstacle avoidance function in two typical non-operational scenarios, namely the narrow environment and the dynamic environment. In the narrow environment, the surroundings are enclosed by fine nets, with several boxes and thin rods in the middle, as shown in Fig. \ref{fig:indoor}(a). In Figure \ref{fig:indoor}(c), the pilot wears the DJI Goggles 2 and specifies the flight targets in real time using the joysticks based purely on the FPV feedback on the Goggle, while the flight safety is assured by the obstacle avoidance function onboard the quadrotor. To enhance the ability to navigate through narrow areas, we set the user-defined obstacle avoidance distance (represented by the blue line in Fig. \ref{fig:indoor}(b)) equal to the quadrotor's size. As shown in Fig. \ref{fig:indoor}(b), the distance $d_{min}$ between the quadrotor’s center and the nearest obstacles remains greater than the quadrotor's size, suggesting a successful obstacle avoidance throughout the whole test. The achievement of obstacle avoidance is primarily attributed to the robustness and effectiveness of the navigation algorithm. The mapping module (Sec. \ref{sec:ref_path}) constructs high-resolution occupancy maps, effectively updating the corresponding grid cells in the map with small objects (\eg, fine nets or thin rods) detected by the LiDAR scans. Additionally, the frontend of the planning and control module (Sec. \ref{sec:ipc}) efficiently generates a collision-free reference path based on joystick commands. Subsequently, the backend employs MPC to achieve high-precision control, ensuring the quadrotor safely arrives at the target point.

\begin{figure}[h]
    \centering
    \includegraphics[width=0.96\textwidth]{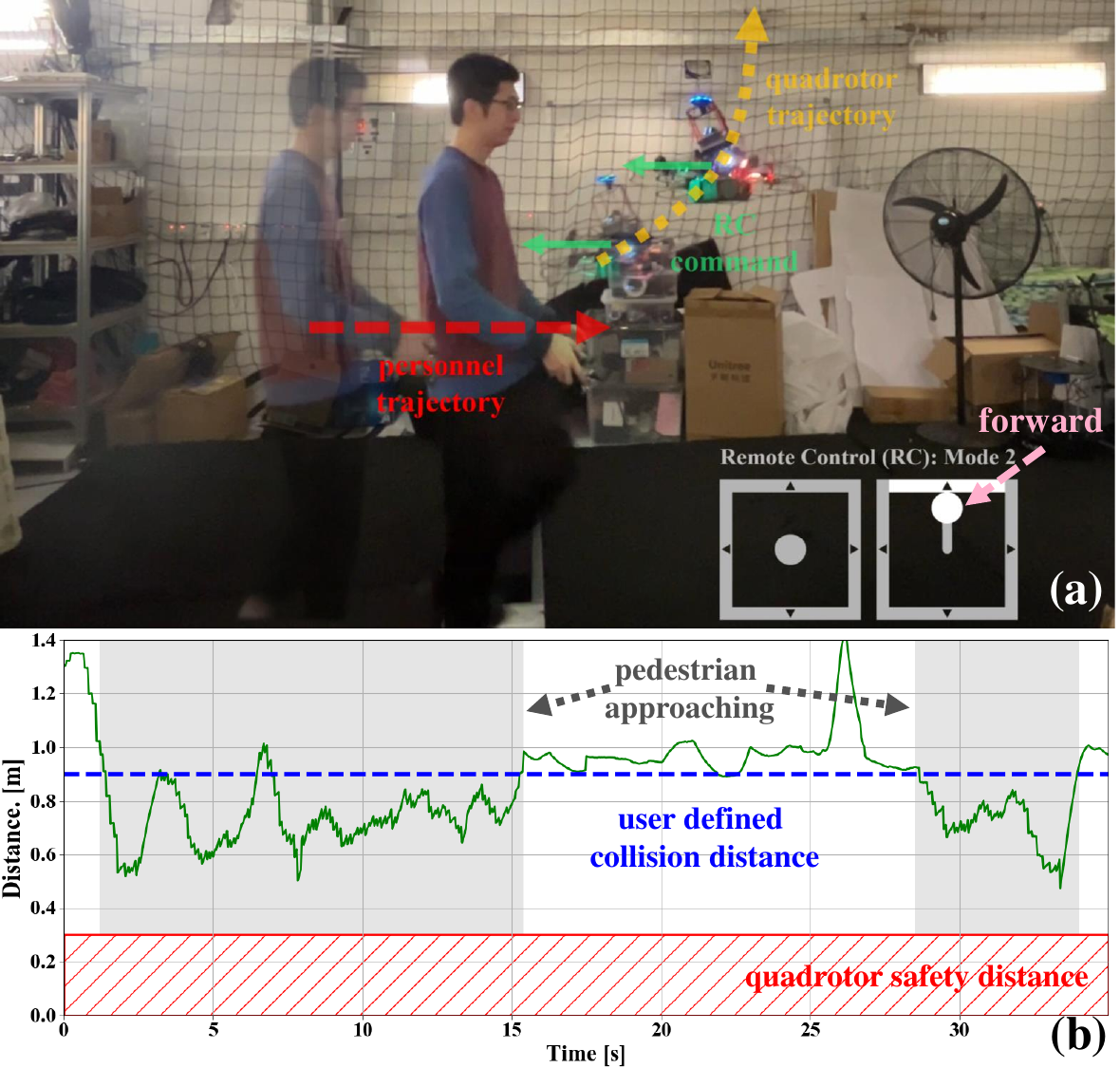}
    \caption{Assisted obstacle avoidance flight in dynamic environment. (a) The third person view of the flight. As the personnel approaches the quadrotor, the quadrotor moves away from the personnel to avoid collisions, despite adverse joystick commands. (b) As the personnel approaches the quadrotor, the distance between the quadrotor's center and the nearest obstacles remains constantly greater than the quadrotor's size throughout the flight.}
    \label{fig:indoor_dyn}
\end{figure}

Next, we evaluate the assisted obstacle avoidance function in dynamic environment, as illustrated in Fig. \ref{fig:indoor_dyn}(a), where the personnel and the quadrotor move towards each other. Under the assisted obstacle avoidance function, the quadrotor ignores the remote control commands directing it towards the personnel and instead maneuvers to avoid him. In this experiment, the user-defined obstacle avoidance distance is set to \SI{0.9}{m} to provide the quadrotor with a larger clearance for reacting and avoiding dynamic objects. As depicted in Fig. \ref{fig:indoor_dyn}(b), although the motion of the dynamic object leads to the distance between the quadrotor and the obstacle being smaller than the user-defined obstacle avoidance distance, the distance remained bigger than the quadrotor safety distance. This is achieved by the low system latency of our navigation algorithm. Upon receiving a LiDAR scan, the mapping module (Sec. \ref{sec:map}) promptly updates the obstacles in the probability and inflated map. Subsequently, the planning and control module, operating at a high frequency with low latency, generates a new local goal in the Known Free region and a reference path reaching the new goal in the frontend (Sec. \ref{sec:ref_path}). This reference path is then leveraged by the backend model predictive control (MPC) (Sec. \ref{sec:backend}) to enable the quadrotor to rapidly respond and navigate towards a safe area. By successfully conducting these two test scenarios, we validate the feasibility of the quadrotor's assisted obstacle avoidance function.

We invite readers to watch our first supplementary video\footnote{\href{https://youtu.be/wqR8NeDTfQU}{https://youtu.be/wqR8NeDTfQU}}, to get a more intuitive understanding of the functional tests of our quadrotor in non-operational scenarios.

\subsection{Six Field Tests}
\label{sec:field_test}

To validate our system's suitability for slope inspection in dense vegetation environments, we collaborate with the Civil Engineering and Development Department (CEDD) to deploy the quadrotor in the field and utilize the onboard camera for close-range visual inspection of the barriers. We conduct tests at six different locations, including five slopes covered with dense vegetation and one slope recently experienced a landslide as a result of the 2023 Hong Kong rainstorm and floods caused by the landfall of Typhoon Haikui\footnote{\href{https://wikipedia.org/wiki/2023_Hong_Kong_rainstorm_and_floods}{https://wikipedia.org/wiki/2023\_Hong\_Kong\_rainstorm\_and\_floods}}. The six field tests are summarized in Table \ref{tab:flight_field}. The third field test, conducted next to the Yiu Hing Road, involves a slope that has just experienced a landslide, where the quadrotor conducts close observation of the barriers and stones in a relatively wide space. The other five field tests consist of slopes with dense vegetation, where the quadrotor performs close-range photographic inspection of the flexible debris-resisting barriers in narrow spaces.

\begin{table}[h]
    \centering
    \small
    \caption{Flight Data from Six Field Tests}
    \label{tab:flight_field}
    \begin{tabular}{|c|c|c|c|c|c|c|}
        \hline
        Test No. & Flexible Barrier No. & Location & Max. Speed & Trajectory & Flight Time \\
         & as referred by CEDD & & $(m/s)$ & Length $(m)$ & $(min : sec)$ \\
        \hline
        1 & 11SW-C/ND3 & Victoria Road, Pokfulam & 1.33 & 133.49 & 7 : 30 \\
        \hline
        2 & 11SW-C/ND6 & Victoria Road, Pokfulam & 1.28 & 79.31 & 3 : 43 \\
        \hline
        3 & - & Yiu Hing Road & 2.82 & 425.99 & 6 : 55 \\
        \hline
        4 & 11SE-B/ND1 & Lei Yue Mun Estate & 1.37 & 274.68 & 8 : 28 \\
        \hline
        5 & 11SE-B/ND2 & Lei Yue Mun Estate & 1.29 & 154.34 & 5 : 24 \\
        \hline
        6(part 1) & 11SW-C/ND11 & Victoria Road, Pokfulam & 1.91 & 186.18 & 8 : 25 \\
        6(part 2) & & & 1.78 & 227.20 & 8 : 28 \\
        \hline
    \end{tabular}
\end{table}

\begin{figure}[h]
    \centering
    \includegraphics[width=0.95\textwidth]{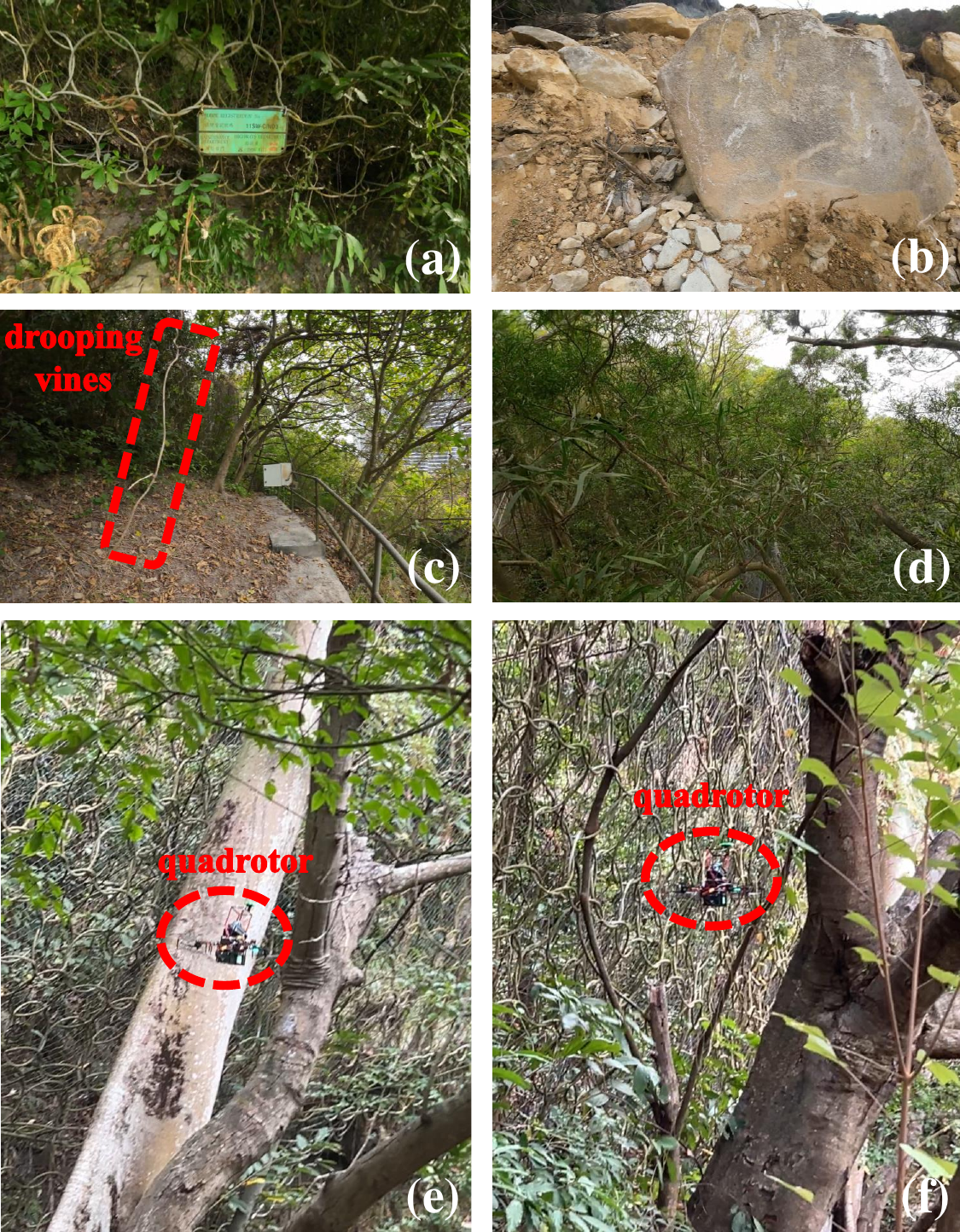}
    \caption{(a) Quadrotor performing close-range inspection of flexible debris-resisting barriers. (b) Quadrotor conducting a close-range inspection of stones caused by landslides. (c) Quadrotor avoiding thin dropping vines. (d) Quadrotor maneuvering through dense tree canopies (first person view). (e)-(f) Quadrotor navigating through narrow tree branches.}
    \label{fig:special_view}
\end{figure}

\begin{figure}[h]
    \centering
    \includegraphics[width=0.93\textwidth]{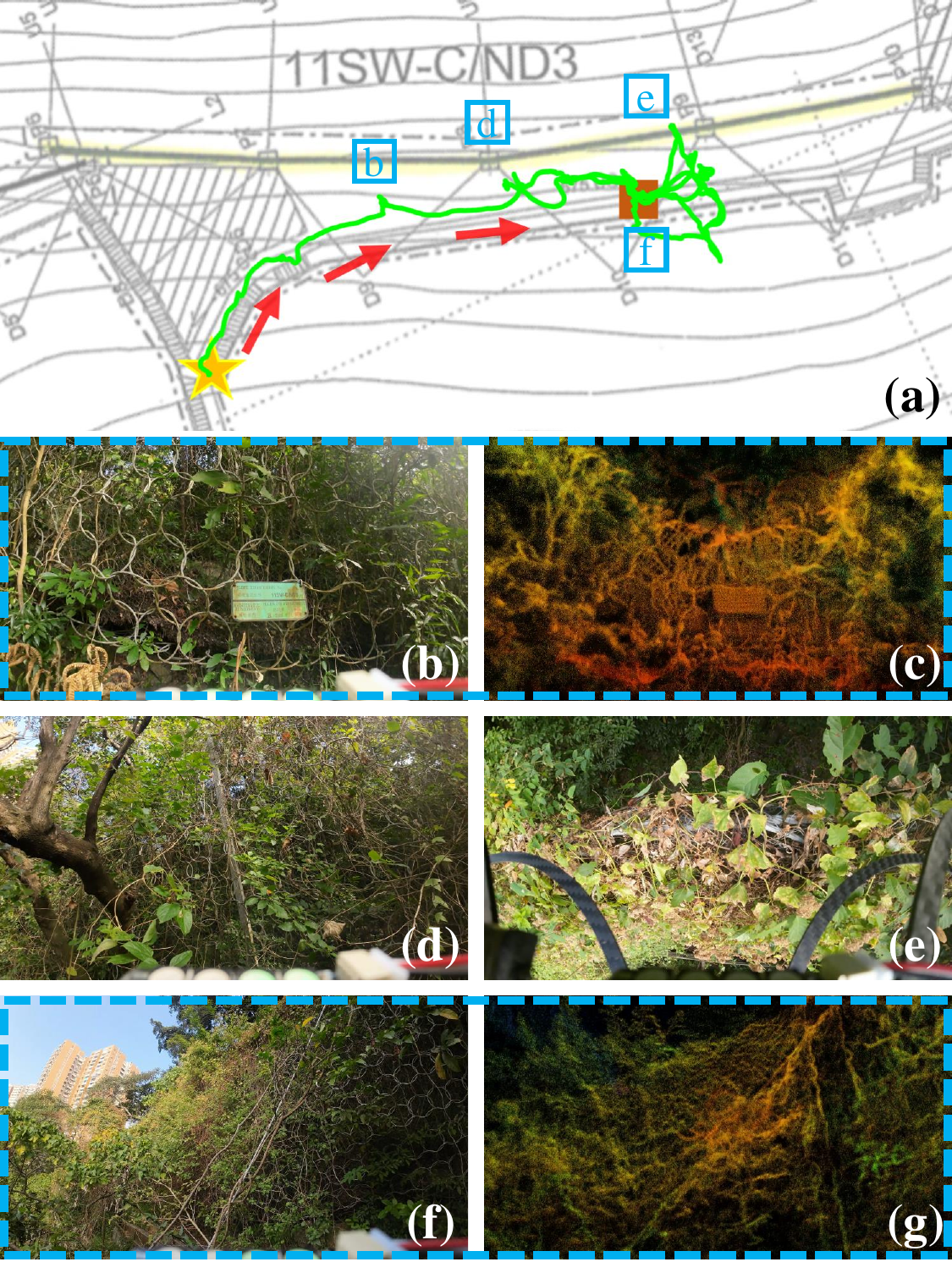}
    \caption{Flight data from the field test of 11SW-C/ND3 slope next to Victoria Road, Pokfulam. (a) The green curve represents the flight trajectory of the quadrotor, the yellow star represents the take-off point, and the brown box represents the landing point. (b), (d), (e) and (f): First-person view photos taken during the inspection. (c) and (g): Point cloud map built from quadrotor's onboard LiDAR.}
    \label{fig:CND3}
\end{figure}

\begin{figure}[h]
    \centering
    \includegraphics[width=\textwidth]{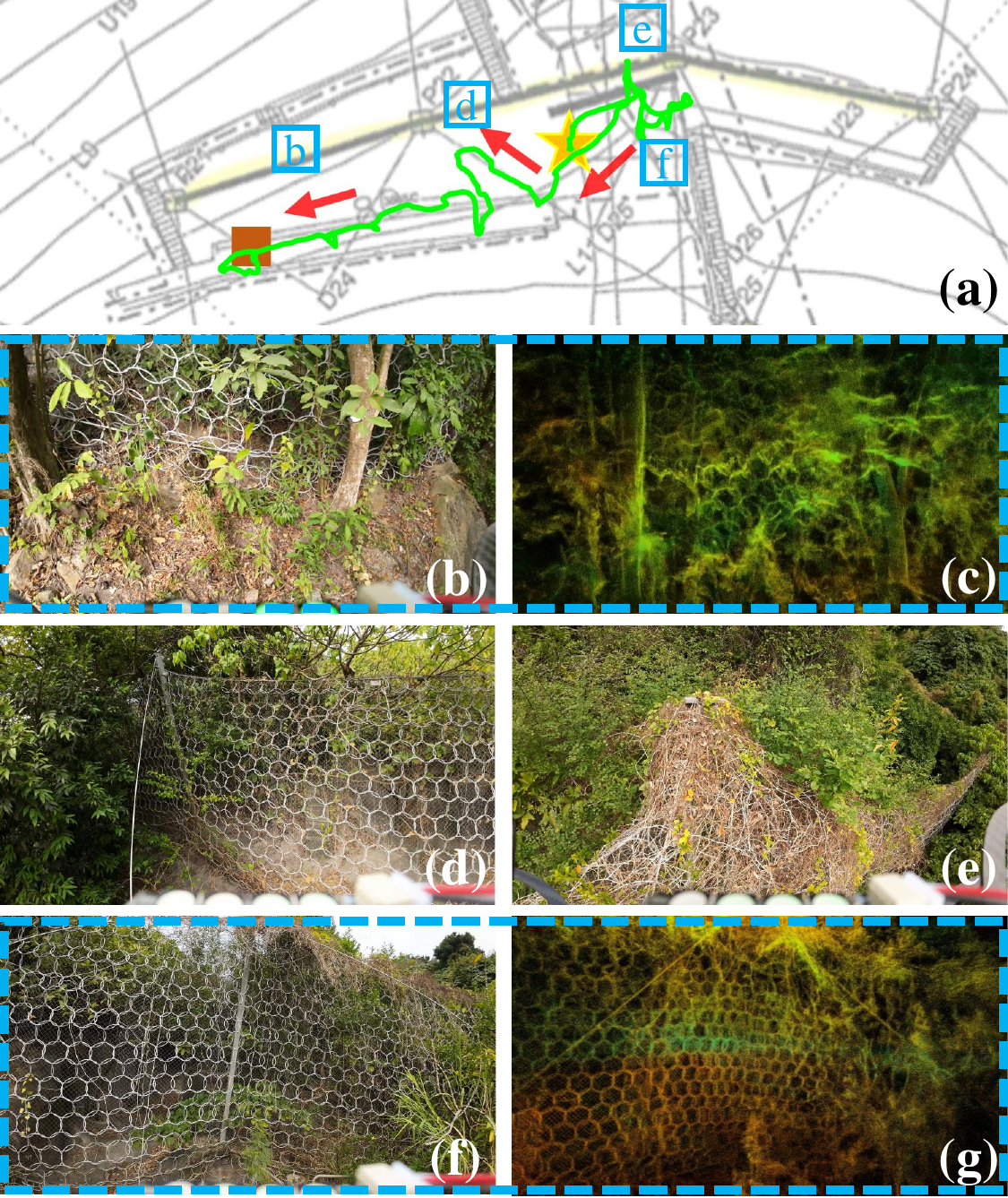}
    \caption{Flight data from the field test of 11SW-C/ND6 slope next to Victoria Road, Pokfulam. (a) The green curve represents the flight trajectory of the quadrotor, the yellow star represents the take-off point, and the brown box represents the landing point. (b), (d), (e) and (f): First-person view photos taken during the inspection. (c) and (g): Point cloud map built from quadrotor's onboard LiDAR.}
    \label{fig:CND6}
\end{figure}

\begin{figure}[h]
    \centering
    \includegraphics[width=0.95\textwidth]{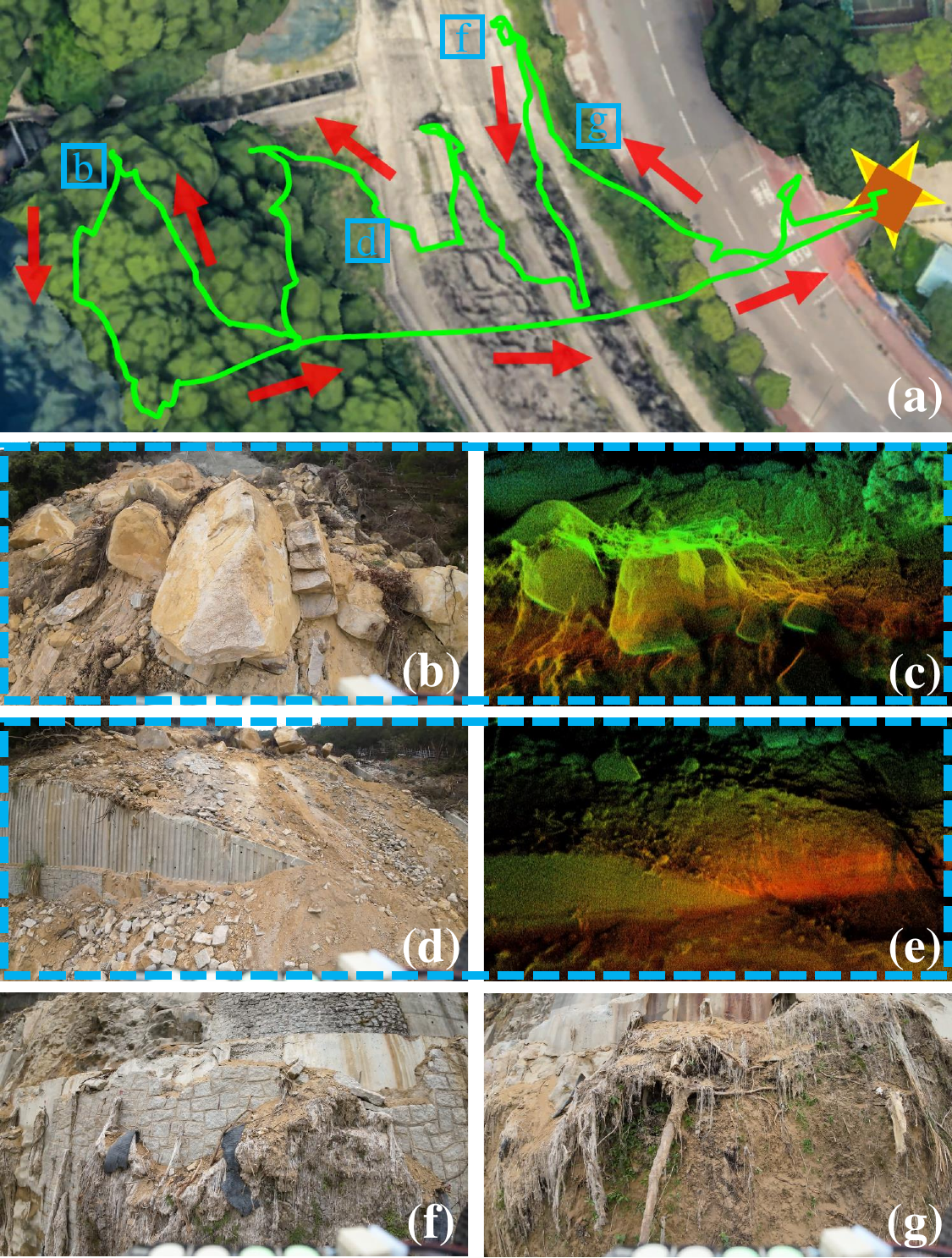}
    \caption{Flight data from the field test of a slope next to Yiu Hing Road. (a) The green curve represents the flight trajectory of the quadrotor, the yellow star represents the take-off point, and the brown box represents the landing point. (b), (d), (f) and (g): First-person view photos taken during the inspection. (c) and (e): Point cloud map built from quadrotor's onboard LiDAR.}
    \label{fig:YiuHingRoad}
\end{figure}

\begin{figure}[h]
    \centering
    \includegraphics[width=\textwidth]{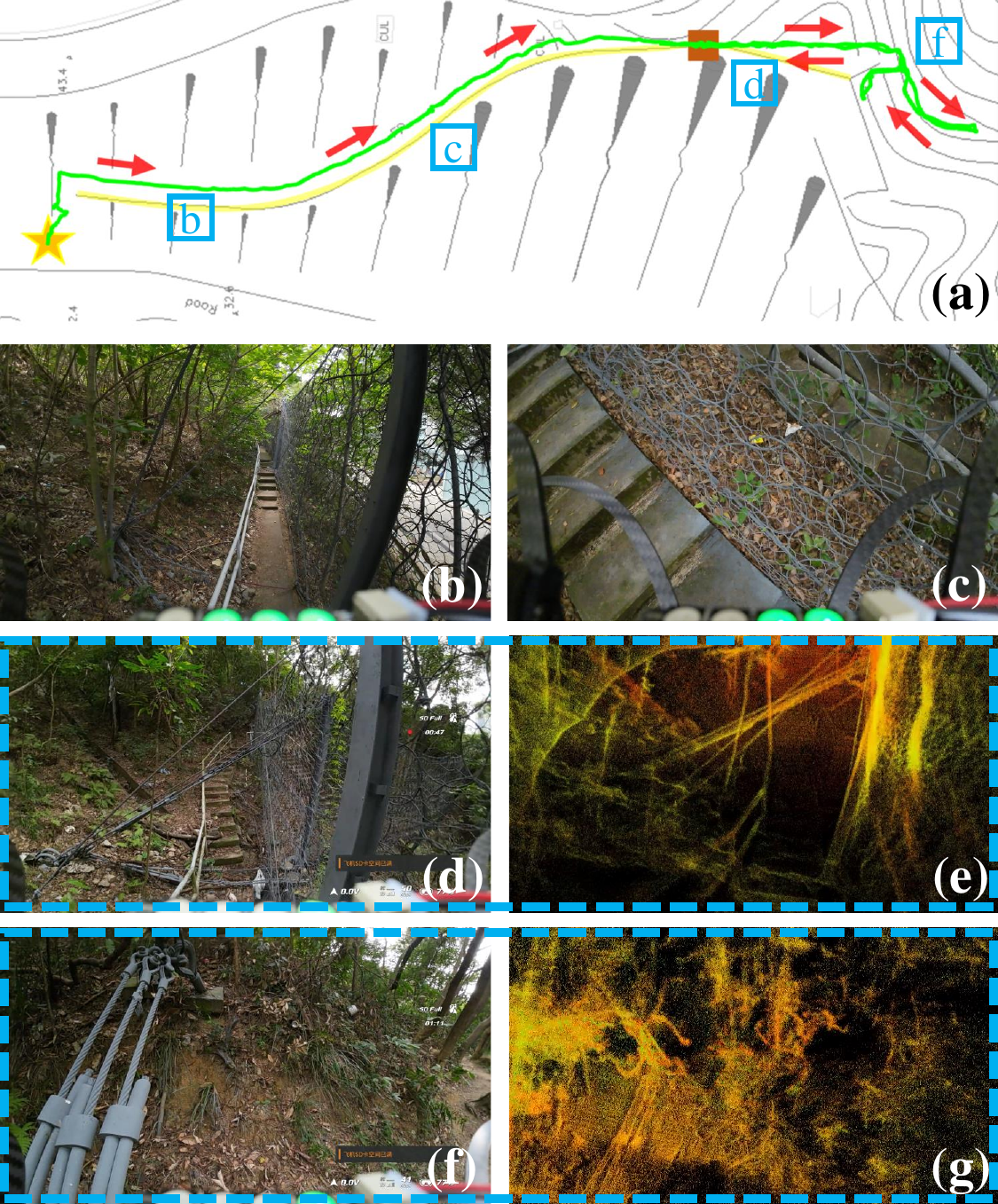}
    \caption{Flight data from the field test of 11SE-B/ND1 slope next to Lei Yue Mun Estate. (a) The green curve represents the flight trajectory of the quadrotor, the yellow star represents the take-off point, and the brown box represents the landing point. (b), (c), (d) and (f): First-person view photos taken during the inspection. (e) and (g): Point cloud map built from quadrotor's onboard LiDAR.}
    \label{fig:BND1}
\end{figure}

\begin{figure}[h]
    \centering
    \includegraphics[width=\textwidth]{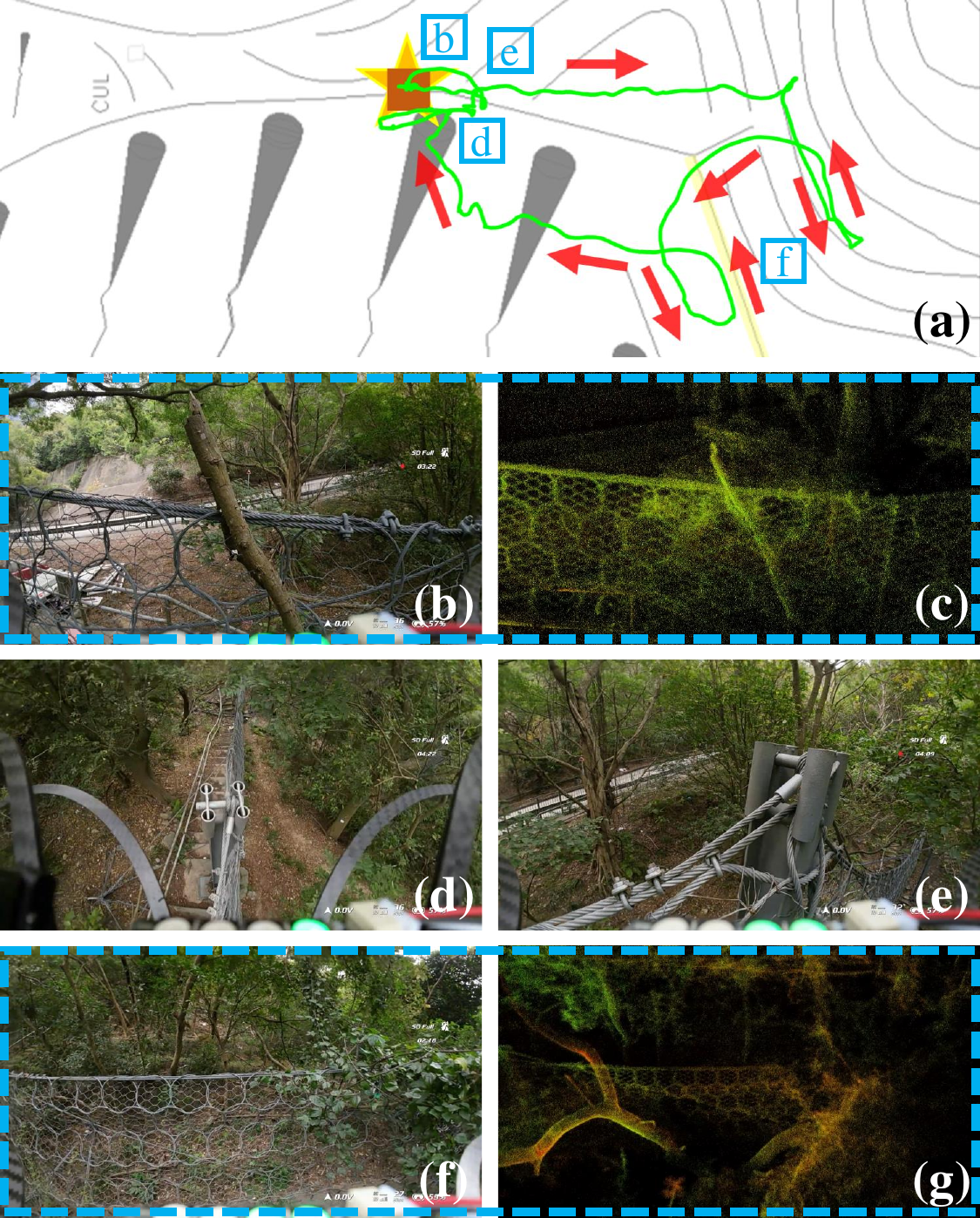}
    \caption{Flight data from the field test of 11SE-B/ND2 slope next to Lei Yue Mun Estate. (a) The green curve represents the flight trajectory of the quadrotor, the yellow star represents the take-off point, and the brown box represents the landing point. (b), (d), (e) and (f): First-person view photos taken during the inspection. (c) and (g): Point cloud map built from quadrotor's onboard LiDAR.}
    \label{fig:BND2}
\end{figure}

\begin{figure}[h]
    \centering
    \includegraphics[width=0.93\textwidth]{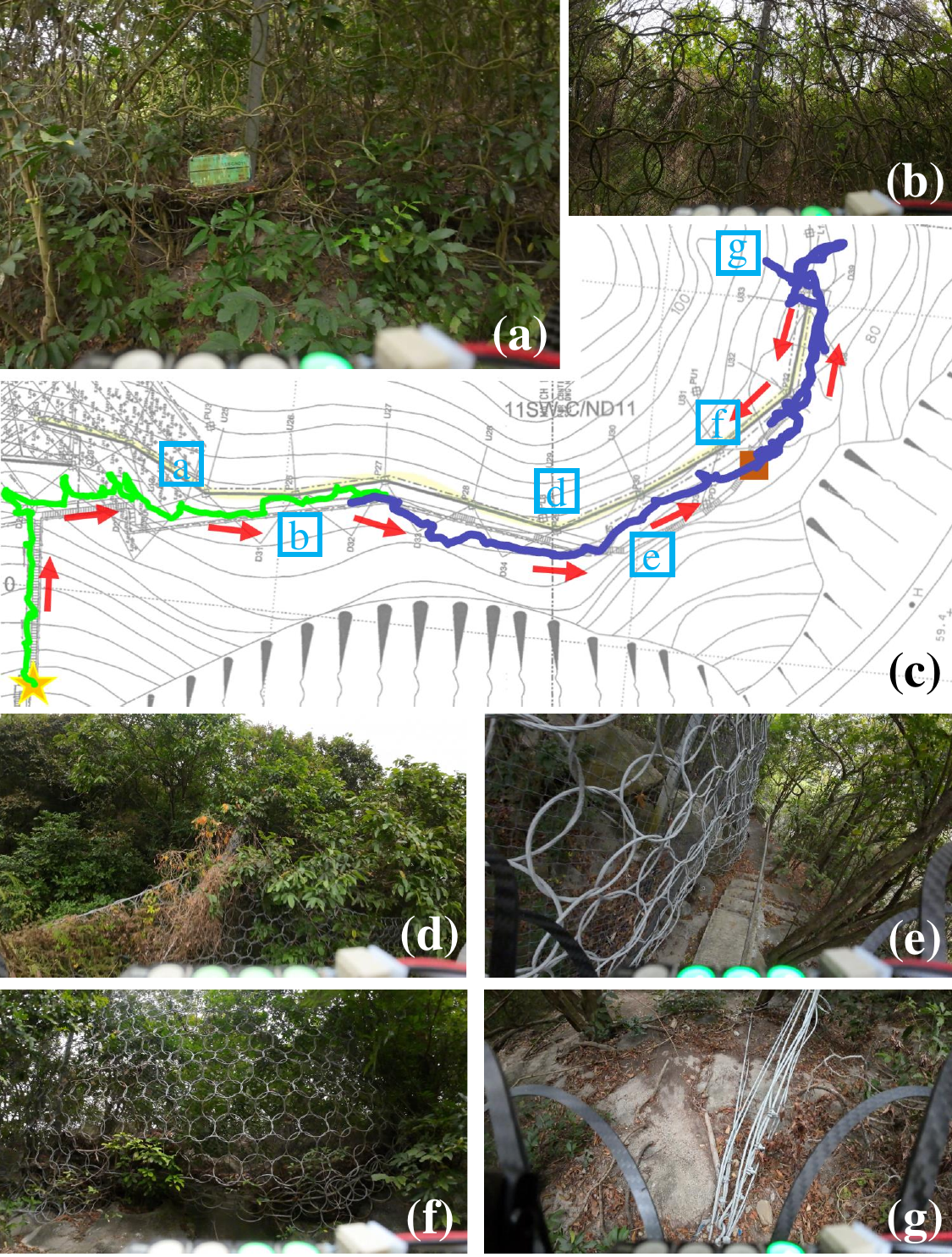}
    \caption{Flight data from the field test of 11SW-C/ND11 slope next to Victoria Road, Pokfulam. (c) The green curve represents the flight trajectory of the first part of the quadrotor, the purple curve represents the flight trajectory of the second part of the quadrotor, the yellow star represents the take-off point, and the brown box represents the landing point. (a), (b) and (d-g): First-person view photos taken during inspection.}
    \label{fig:CND11}
\end{figure}

With the assistance of obstacle avoidance functionality, our quadrotor assists the pilot in close-range visual inspection of flexible debris-resisting barriers and stones resulting from landslides (Fig. \ref{fig:special_view}(a) and Fig. \ref{fig:special_view}(b)), maneuvering through dense tree canopies (Fig. \ref{fig:special_view}(c)), avoiding thin dropping vines (Fig. \ref{fig:special_view}(d)), and navigating through narrow tree branches (Fig. \ref{fig:special_view}(e) and Fig. \ref{fig:special_view}(f)). Ultimately, our quadrotor completes all six field tests, demonstrating its suitability for slope inspection in dense vegetation environments. The flight trajectory of each test is superimposed on the slope drawing provided by the CEDD, the point cloud map built online by our navigation system, and the example photos taken during the inspection are shown in Fig. \ref{fig:CND3} to \ref{fig:CND11}. These outputs provide the Hong Kong CEDD with detailed and valuable information about the inspected areas, enabling thorough analysis and assessment of the flexible debris-resisting barriers' condition.

Throughout these six field tests, the quadrotor benefited from the accurate perception of thin objects provided by the LiDAR sensor. This accurate perception allows our quadrotor to construct high-resolution local occupancy grid maps in the mapping module (Sec. \ref{sec:map}) and update the occupancy status of grid cells containing moving branches and other thin objects. Moreover, thanks to the assisted obstacle avoidance function in the frontend of the planning and control module (Sec. \ref{sec:ref_path}), our quadrotor is capable of safely approaching inspection targets, such as flexible debris-resisting barriers, at distances as small as the quadrotor's size. This enables the quadrotor to capture high-definition images for subsequent detailed analysis. Additionally, despite encountering varying degrees of natural wind disturbances, the quadrotor effectively suppresses these disturbances without compromising flight performance, due to the real-time generation of optimal control actions by the model predictive control (MPC) problem in the IPC's backend (Sec. \ref{sec:backend}).

To gain a more comprehensive understanding of our quadrotor's real-world performance in the field tests, we invite readers to watch our second supplementary video\footnote{\href{https://youtu.be/Uy3yYAmmeM0}{https://youtu.be/Uy3yYAmmeM0}}\footnote{\href{https://youtu.be/mTmR8C3OVkI}{https://youtu.be/mTmR8C3OVkI}}. The video provides a detailed showcase of our quadrotor's performance during the field test of the 11SW-C/DN11 slope next to Victoria Road, Pokfulam. Moreover, we present first-person videos recorded during the other five field tests.

\subsection{Benchmark with DJI Mavic 3}
\label{sec:benchmark}

\begin{figure}[h]
    \centering
    \includegraphics[width=\textwidth]{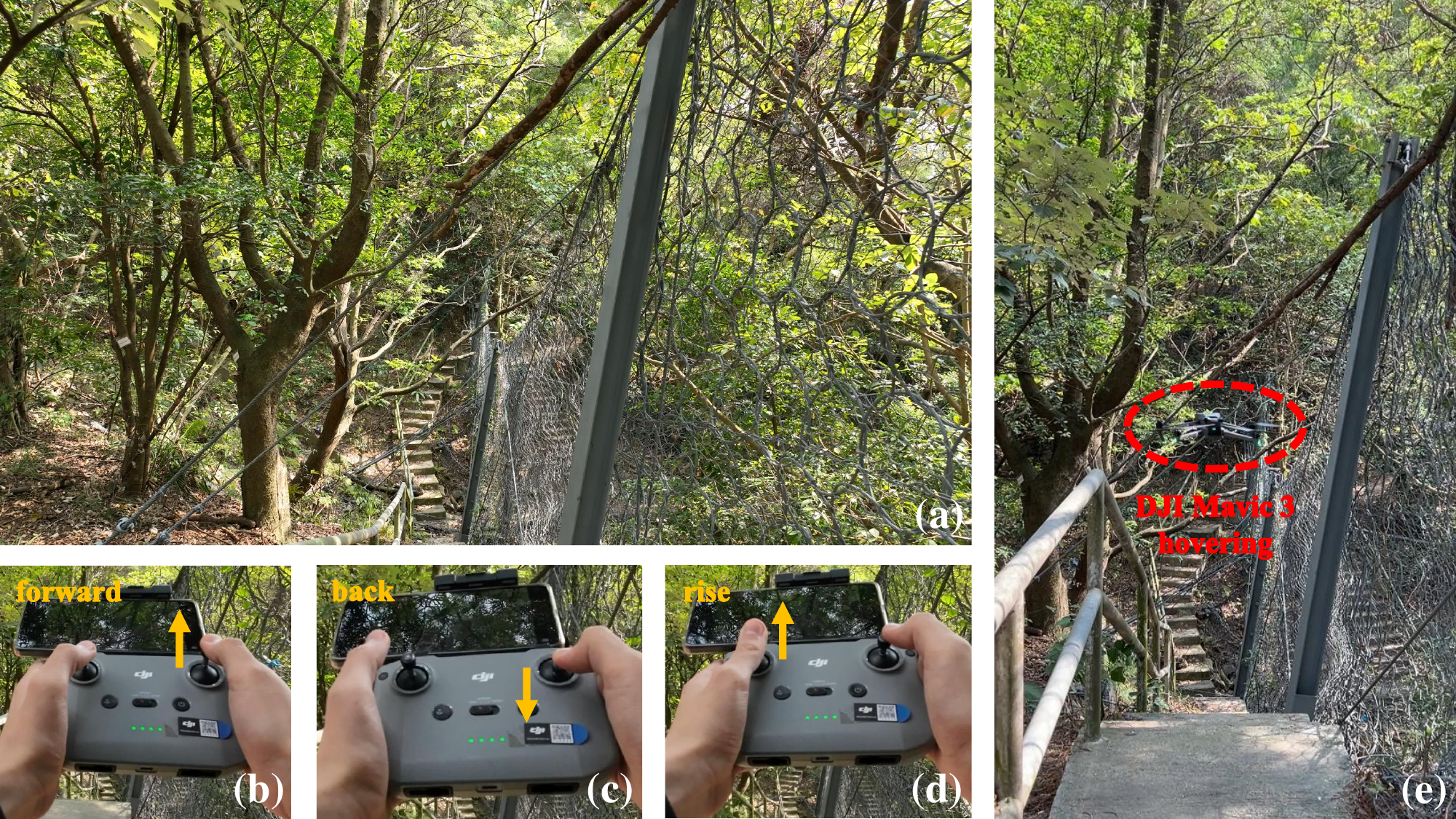}
    \caption{DJI Mavic 3 exhibiting conservative behavior in Normal Bypass mode. Similar behavior is observed in Brake mode within the same environment. (a) First-person view captured from the DJI Mavic 3. (b)-(d) Remote Control commands issued by the pilot. (e) Third-person view of the DJI Mavic 3. Despite a clear feasible corridor in the commanded direction, the DJI Mavic 3 fails to follow the pilot's commands and remains hovering in place.}
    \label{fig:dji1}
\end{figure}

DJI Mavic 3 \cite{mavic3}, as one of the most advanced commercial drones, is equipped with up to eight wide-angle cameras and incorporates the advanced pilot assistance system (APAS 5.0) algorithm for high-level flight assistance, enabling omnidirectional obstacle sensing. It offers three obstacle avoidance modes: Brake, Normal Bypass, and Nifty Bypass, allowing pilots to customize the settings based on the environments and their preferences. In Brake mode, the DJI Mavic 3 comes to an immediate stop if in the flight direction an obstacle is detected. In Normal Bypass and Nifty Bypass modes, the DJI Mavic 3 can bypass obstacles, but in Nifty Bypass mode, it maintains a smaller clearance with obstacles, so possessing a higher passability but also a higher risk of collision. In sum, among these three modes, the Brake mode provides the highest level of flight safety, followed by the Normal Bypass mode, and the Nifty Bypass mode performs the least effectively. However, in terms of accessibility in narrow areas, the order is reversed, with Nifty Bypass mode performing the best, Normal Bypass mode being in the middle, and Brake mode performing the worst.

\begin{figure}[h]
    \centering
    \includegraphics[width=\textwidth]{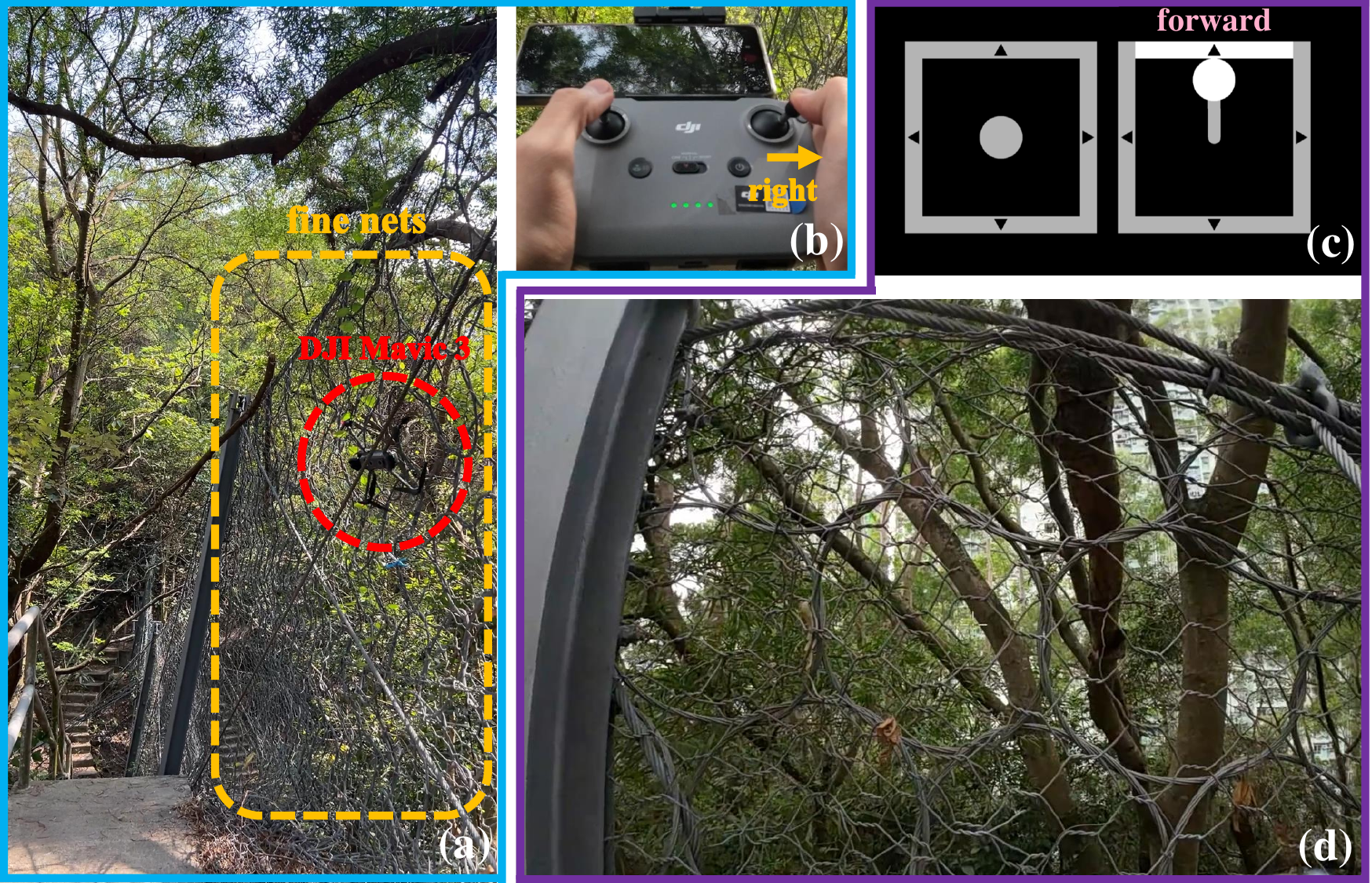}
    \caption{(a) DJI Mavic 3 encountering collisions with fine nets in Brake mode. (b) Remote Control commands issued by the pilot during DJI Mavic 3's collision with fine nets. (c) Joystick commands commanding our quadrotor to fly towards the fine nets. (d) Our quadrotor perceives the fine nets ahead and gives up executing the joystick commands by stops in front of the nets.}
    \label{fig:dji2}
\end{figure}

To further validate the suitability of our quadrotor for slope inspection in dense vegetation, we conduct tests on the 11SE-B/ND1 slope next to Lei Yue Mun Estate (Table. \ref{tab:flight_field}) to compare its obstacle avoidance function with DJI Mavic 3 in different modes. When operating in Normal Bypass modes, DJI Mavic 3 exhibits highly conservative behavior in dense vegetation environments, as shown in Fig. \ref{fig:dji1}. In such an environment, despite a clear feasible corridor in the commanded direction, the DJI Mavic 3 prioritizes safety by maintaining a hover in place. The Brake mode behaves similarly to the Normal Bypass mode, as it remains hovering too. While the Brake mode prioritizes the most on safety thus being highly conservative, our testing revealed that when flying towards fine nets, it failed to execute the necessary stop maneuver, resulting in a collision, as shown in Fig. \ref{fig:dji2}(a). This deficiency can be attributed to its limited perception capabilities, particularly when dealing with thin objects. In contrast, as depicted in Fig \ref{fig:dji2}(c) and Fig \ref{fig:dji2}(d), our LiDAR-based quadrotor effectively perceives the presence of fine nets ahead, actively refusing to follow joystick commands to fly towards them, thus ensuring flight safety.

\begin{figure}[h]
    \centering
    \includegraphics[width=\textwidth]{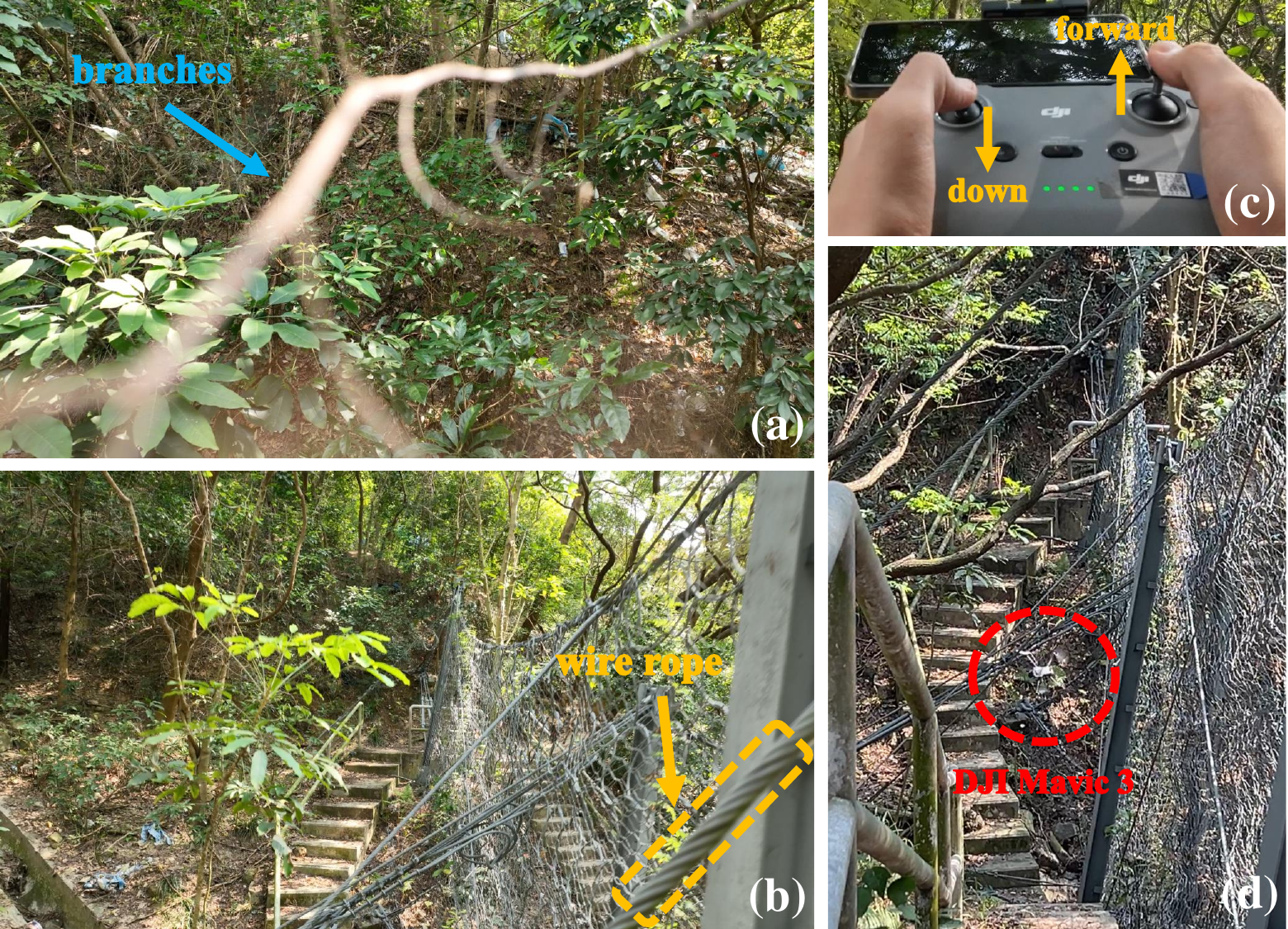}
    \caption{(a) DJI Mavic 3 experiencing collisions with tree branches in Nifty Bypass mode. (b) DJI Mavic 3 colliding with a wire rope in Nifty Bypass mode. (c) The Remote Control commands issued by the pilot when DJI Mavic 3 collides with the wire rope. (d) DJI Mavic 3 colliding with a wire rope when operating in Nifty Bypass mode.}
    \label{fig:dji3}
\end{figure}

In Nifty Bypass mode, DJI Mavic 3 can fly in relatively open areas on maintenance access. However, it should be noted that the flight safety in this mode is the worst, as the rotor blades are prone to collide with tree leaves and thin branches, as shown in Fig. \ref{fig:dji3}(a). Besides, DJI Mavic 3 can easily give the pilot a false feeling of loss of control. For example, it often refuses to follow the pilot's commands when in proximity to the inspection targets, while exhibits unexpected large maneuvers and long flight distances when otherwise. Moreover, despite Nifty Bypass mode being the most aggressive, the pilot still lacks control when navigating narrow areas where the available space is less than twice the size of the DJI Mavic 3. Moreover, DJI Mavic 3 is even more susceptible to collisions and crashes while operating in this mode. As illustrated in Fig. \ref{fig:dji3}(c), when directed by the pilot to fly in the forward downward region, it fails to detect and avoid the wire rope shown in Fig. \ref{fig:dji3}(b), resulting in a collision depicted in Fig. \ref{fig:dji3}(d).

% Besides, when executing pilot commands, DJI Mavic 3 performs large maneuvers and travels significant distances, which can give the pilot a false feeling of loss of control. 

\begin{figure}[h]
    \centering
    \includegraphics[width=\textwidth]{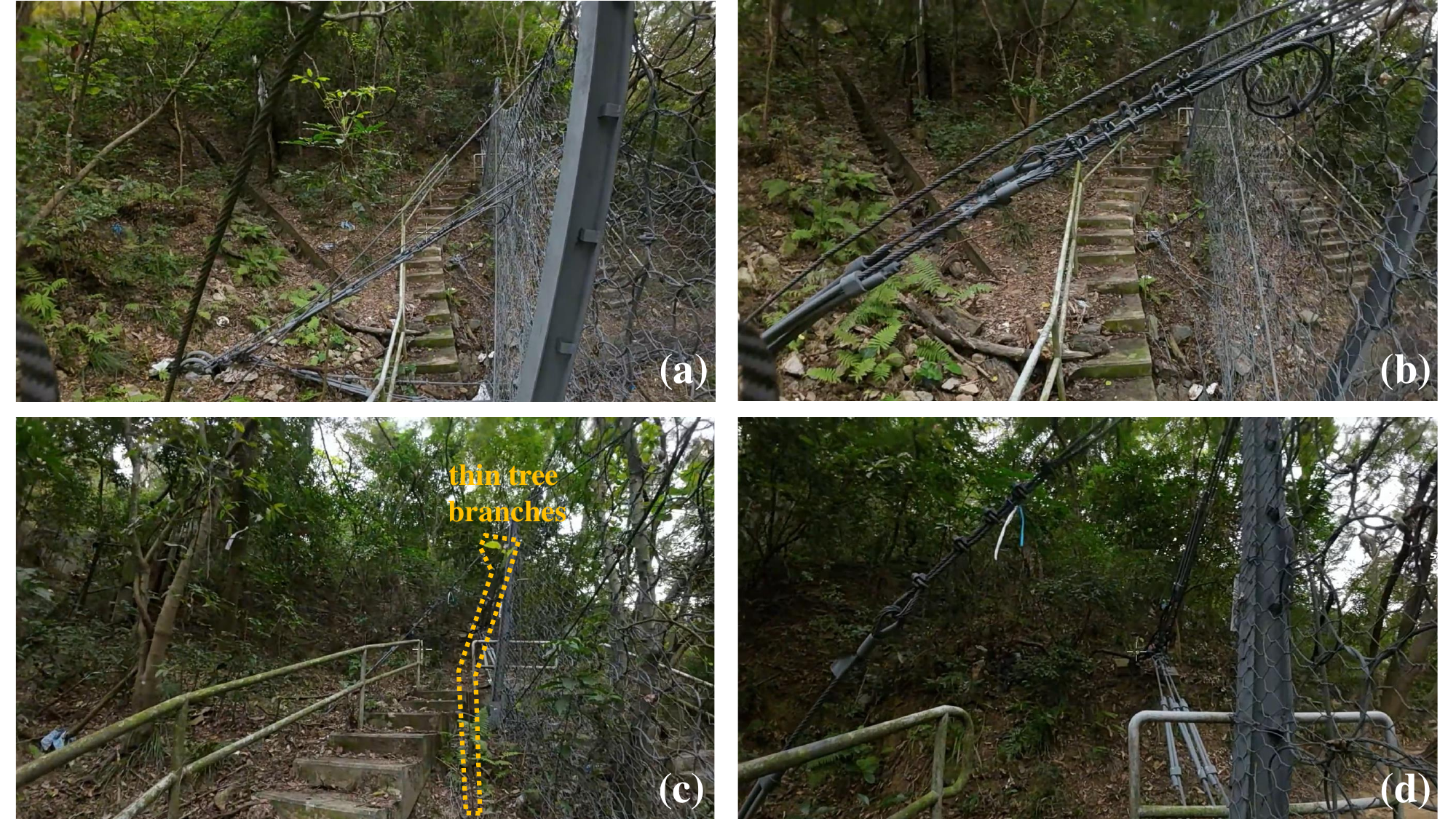}
    \caption{First-person view pictures taken by our quadrotor when inspecting the 11SE-B/ND1 slope next to Lei Yue Mun Estate.}
    \label{fig:bench_uav}
\end{figure}

Overall, in dense vegetation environments, the Brake mode and Normal Bypass mode of the DJI Mavic 3 prove to have insufficient ability to navigate through dense crowded vegetation environments and encounter frequent immediate stops in the presence of clear flight passage. Moreover, even in the most conservative Brake mode, the DJI Mavic 3 fails to perceive fine nets, resulting in collisions. The Nifty Bypass mode of the DJI Mavic 3 has improved passability in dense vegetation, but at the cost of much lower safety level, often leading to collisions with small objects, such as tree leaves, tree branches, and wire ropes. On the other hand, our LiDAR-based quadrotor performs exceptionally well in the same test scenario at the 11SE-B/ND1. As shown in Fig. \ref{fig:bench_uav}, our quadrotor maneuvers agilely through narrow areas while still roughly following the pilot's commands. Furthermore, our quadrotor effectively avoids collisions with small objects such as thin tree branches (Fig. \ref{fig:bench_uav}(c)), successfully executing the necessary stop maneuvers when encountering fine nets (Fig. \ref{fig:dji2}(d)).

\begin{figure}[h]
    \centering
    \includegraphics[width=\textwidth]{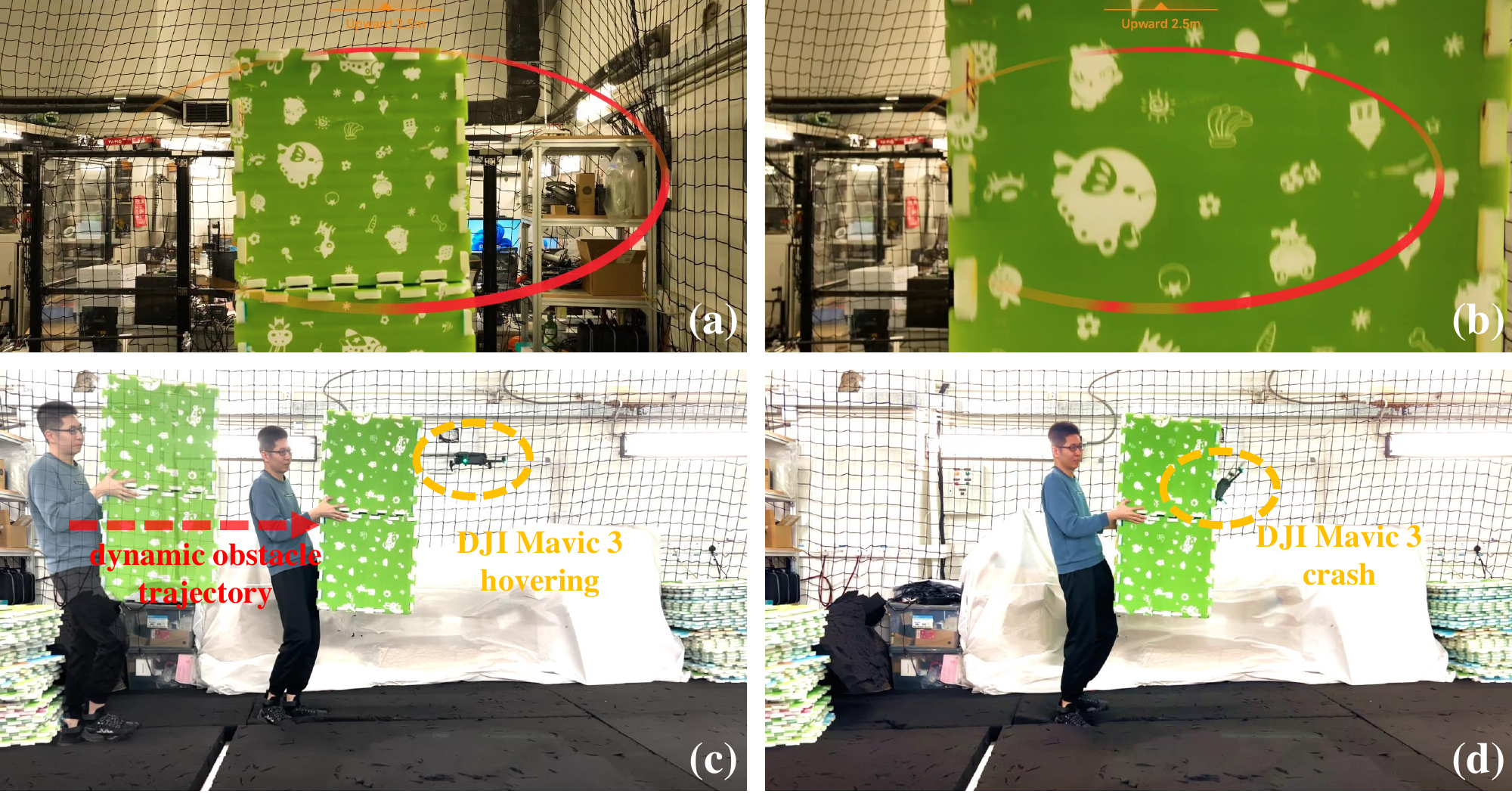}
    \caption{In Nifty Bypass mode, DJI Mavic 3, despite its ability to perceive dynamic objects, cannot avoid them, resulting in crashes. (a)-(b) First-person view of DJI Mavic 3 as a dynamic obstacle approaches. (c) Slow approach of a dynamic obstacle, while the DJI Mavic 3 remains hovering in place. (d) DJI Mavic 3 fails to avoid the dynamic object and results in a crash.}
    \label{fig:dji_dyn}
\end{figure}

Finally, in terms of dynamic obstacle avoidance, as shown in Fig. \ref{fig:dji_dyn}, in Nifty Bypass mode, the DJI Mavic 3 is able to perceive dynamic obstacle ahead but fails to avoid it, further reducing its safety assurance. In comparison, our quadrotor can successfully evade slow-moving dynamic objects, as shown in Fig. \ref{fig:indoor_dyn}(a). The complete comparison experiments between our quadrotor and the DJI Mavic 3 can be found in the third supplementary video\footnote{\href{https://youtu.be/jTrrS4-O4xY}{https://youtu.be/jTrrS4-O4xY}}.

\section{Conclusion}
\label{sec:conclusion}

In this work, we developed a LiDAR-based quadrotor system specifically designed for slope inspection in dense vegetation. In terms of hardware structure, our quadrotor is equipped with a LiDAR sensor and a high-resolution camera, enabling the collection of photos and point cloud data of the terrain and inspection targets. Its compact size allows it to navigate through narrow areas, and it has a flight time of 12 minutes. On the software side, we developed a comprehensive suite of navigation algorithms specifically tailored to address the challenges posed by dense vegetation environments. These algorithms encompass localization, mapping, planning, and control, enabling the quadrotor to perform assisted obstacle avoidance flight, close-range imaging, and three-dimensional point cloud reconstruction in narrow spaces. The key focus of our navigation algorithms lies in mapping, planning and control. In the mapping module, we incorporated three enhancements to our previous work ROG-Map, including Unknown Grid Cells Inflation, Infinite Points Ray Casting and Incremental Frontiers Update. Unknown Grid Cells Inflation expands the unknown areas to avoid potential collisions with obstacles in the unknown area, providing a higher level of safety assurance. Infinite Points Ray Casting tackles the issue of no LiDAR returned points when facing the sky. Incremental Frontiers Update efficiently updates frontier information based on the latest sensor data. In the planning and control module, we redesigned the frontend and backend of our previous work IPC, to enable assisted obstacle avoidance flight based on the pilot's joystick signals.

To validate the feasibility of our solution, we first conducted functional tests in non-operational scenarios. Subsequently, our quadrotor was deployed in real-world environments and completed six field tests. Additionally, we conducted benchmark experiments between our quadrotor and DJI Mavic 3 to further highlight the advantages of our quadrotor in narrow area flight and dynamic obstacle avoidance. Through these experiments, we demonstrated the suitability of our quadrotor for slope inspection in dense vegetation.

\subsubsection*{Acknowledgments}

This paper is published with the permission of the Head of the Geotechnical Engineering Office and the Director of Civil Engineering and Development, the Government of the Hong Kong Special Administrative Region. This work was supported by the Hong Kong Civil Engineering and Development Department (CEDD). The authors would like to thank Mr. Sammy Cheung from the Hong Kong CEDD for his support, insightful discussions and valuable suggestions on improving the writing of this article. His expertise greatly contributed to the quality and clarity of our work. The authors would also like to express their gratitude to Mr. Ruize Xue, Mr. Bowen Wang, and Ms. Minghe Chen for their valuable assistance during the experiments. The authors gratefully acknowledge DJI for fund support and Livox Technology for equipment support during the project.

\bibliographystyle{apalike}
\bibliography{jfrExampleRefs}

\end{document}